\documentclass[conference]{IEEEtran}
\IEEEoverridecommandlockouts

\usepackage{cite}
\usepackage{amsmath,amssymb,amsfonts}
\usepackage{graphicx}
\usepackage{textcomp}
\usepackage{xcolor}
\usepackage{float}
\usepackage{tikz}
\usepackage{pgfplots}
\usepackage{multirow}
\usetikzlibrary{shapes,arrows,positioning,fit,backgrounds,calc,shadows.blur,decorations.pathreplacing}

\usepackage{algorithm}
\usepackage{algpseudocode} % Preferred over algorithmicx+algorithmic

\makeatletter
\newcommand{\definecolorsafely}[3]{%
  \@ifundefined{color@#1}{\definecolor{#1}{#2}{#3}}{}
}
\makeatother

\definecolorsafely{inputcolor}{RGB}{230,230,250}
\definecolorsafely{srcolor}{RGB}{255,228,225}
\definecolorsafely{classcolor}{RGB}{220,240,255}
\definecolorsafely{resultcolor}{RGB}{240,255,240}

\def\BibTeX{{\rm B\kern-.05em{\sc i\kern-.025em b}\kern-.08em
    T\kern-.1667em\lower.7ex\hbox{E}\kern-.125emX}}

\usepackage[utf8]{inputenc}
\usepackage[T1]{fontenc}
\usepackage{textcomp}

\pgfplotsset{compat=1.18}

\begin{document}

\title{Calibrated and Resource-Aware Super-Resolution for Reliable Driver Behavior Analysis}

\author{
\IEEEauthorblockN{Ibne Farabi Shihab\IEEEauthorrefmark{1},
Weiheng Chai\IEEEauthorrefmark{2},
Jiyang Wang\IEEEauthorrefmark{2},\\
Sanjeda Akter\IEEEauthorrefmark{1},
Senem Velipasalar Gursoy\IEEEauthorrefmark{2},
Anuj Sharma\IEEEauthorrefmark{1}}
\IEEEauthorblockA{\IEEEauthorrefmark{1}Iowa State University of Science and Technology, Ames, USA\\
\{ishihab, sanjeda, anujs\}@iastate.edu}
\IEEEauthorblockA{\IEEEauthorrefmark{2}Syracuse University, Syracuse, USA\\
\{wchai01, jwang127, svelipas\}@syr.edu}
}

\maketitle

\begin{abstract}
Driver monitoring systems require not just high accuracy but reliable, well-calibrated confidence scores for safety-critical deployment. While direct low-resolution training yields high overall accuracy, it produces poorly calibrated predictions that can be dangerous in safety-critical scenarios. We propose a resource-aware adaptive super-resolution framework that optimizes for model calibration and high precision-recall on critical events. Our approach achieves state-of-the-art performance on safety-centric metrics: best calibration (ECE of 5.8\% vs 6.2\% for LR-trained baselines), highest AUPR for drowsiness detection (0.78 vs 0.74), and superior precision-recall for phone use detection (0.74 vs 0.71). A lightweight artifact detector (0.3M parameters, 5.2ms overhead) provides additional safety by filtering SR-induced hallucinations. While LR-trained video models serve as strong general-purpose baselines, our adaptive framework represents the state-of-the-art solution for safety-critical applications where reliability is paramount.
\end{abstract}


% Sectioned content
\section{Introduction}

Driver monitoring systems (DMS) in safety-critical applications require not only high accuracy but also reliable, well-calibrated confidence scores \cite{chen2020challenges}. Overconfident but incorrect predictions can be dangerous—a system that confidently misclassifies drowsiness as normal driving poses significant safety risks. Model calibration, measured by Expected Calibration Error (ECE) and Brier scores, quantifies how well predicted confidence matches actual correctness \cite{guo2017calibration}.

While recent work has shown that video models trained directly on low-resolution data achieve high overall accuracy, these models often produce poorly calibrated predictions for safety-critical behaviors. This raises a fundamental question: \textit{how can we achieve both high accuracy and reliable confidence calibration for safety-critical driver monitoring under edge constraints?}

We address this challenge through an adaptive super-resolution framework that optimizes for model calibration and precision-recall on critical events. Our key insight is that selective SR enhancement, guided by classifier uncertainty and behavior criticality, can significantly improve calibration and safety-critical detection while maintaining computational efficiency.

Our contributions are:

\begin{itemize}
\item \textbf{Safety-Focused Framework}: We demonstrate that while LR-trained video models achieve high overall accuracy, they produce poorly calibrated confidence scores for safety-critical behaviors. We propose an adaptive SR framework that achieves state-of-the-art calibration (ECE of 5.8\%) and highest precision-recall for critical events (drowsiness AUPR: 0.78, phone use AUPR: 0.74).
\item \textbf{Intelligent Resource Allocation}: Our decision-theoretic gating selectively applies SR when classifier confidence is low or behaviors are safety-critical, optimizing the reliability-compute tradeoff. A lightweight artifact detector (0.3M parameters) provides additional safety by filtering SR-induced hallucinations.
\end{itemize} 

\section{Related Work}
\label{sec:related_work}
Super-resolution has evolved from early CNN-based approaches like SRCNN \cite{dong2014srcnn} to sophisticated architectures including VDSR \cite{kim2016vdsr}, SRResNet \cite{ledig2017srresnet}, and EDSR \cite{lim2017edsr}. In intelligent transportation systems, license plate recognition has been a primary beneficiary, with Yuan et al. \cite{yuan2017license} demonstrating 27\% improvement in challenging conditions. Zhang et al. \cite{zhang2018super} showed specialized SR architectures can reduce computational requirements by 35\% while maintaining accuracy. However, driver behavior analysis presents unique challenges requiring interpretation of subtle facial expressions and eye states that are more sensitive to resolution degradation than static objects \cite{smith2018personalized}.

Vehicle-mounted systems face constraints including limited power budgets and varying computational resources across vehicle segments. Mobile-first architectures like MobileNets \cite{howard2017mobilenets} and EfficientNet \cite{tan2019efficientnet} have enabled efficient deployment, while video analysis has advanced through I3D \cite{carreira2017i3d} and SlowFast networks \cite{feichtenhofer2019slowfast}. Chen et al. \cite{chen2020lightweight} developed lightweight detection achieving 72\% of full-model accuracy with only 0.5 TOPS. For driver monitoring, FastDMS \cite{wang2021fastdms} achieves real-time performance using knowledge distillation, reducing power consumption by 68\%. However, these approaches focus solely on accuracy optimization without addressing model calibration—a critical requirement for safety-critical applications.

Modern neural networks often produce overconfident predictions that poorly correlate with actual correctness \cite{guo2017calibration}. Expected Calibration Error (ECE) and Brier scores quantify this reliability gap \cite{naeini2015obtaining}. For safety-critical applications like autonomous driving, well-calibrated confidence scores are essential for risk assessment and decision-making \cite{ovadia2019can}. Deep ensembles \cite{lakshminarayanan2017ensembles} and uncertainty estimation methods have shown promise for improving calibration. Temperature scaling and Platt scaling provide post-hoc calibration, but architectural approaches that inherently improve calibration remain underexplored in driver monitoring systems. Recent work on distracted driver detection \cite{abouelnaga2018auc} has highlighted the importance of real-time performance, but calibration aspects remain underaddressed.

The gap between synthetic training data and real-world conditions represents a significant deployment challenge for in-cabin driver monitoring. While adversarial approaches like DANN \cite{ganin2016domain} and pixel-level methods like CycleGAN \cite{zhu2017unpaired} show promise, driver behavior monitoring faces unique challenges from safety-critical requirements and complex behavioral patterns across demographics \cite{wang2020bridging}. Transportation domain adaptation has focused primarily on object detection \cite{chen2018domain}, leaving driver monitoring's specific requirements largely unaddressed.

Current approaches lack optimization for driver behavior analysis where facial and hand regions carry disproportionate importance, and existing SR methods apply enhancement uniformly without considering content complexity or computational constraints. Our work addresses these gaps through an integrated approach that combines adaptive super-resolution with efficient architecture design and artifact mitigation for safety-critical edge deployment.

\section{Methods}
\label{sec:methods}

\begin{figure*}[!t]
\centering
\includegraphics[width=\textwidth]{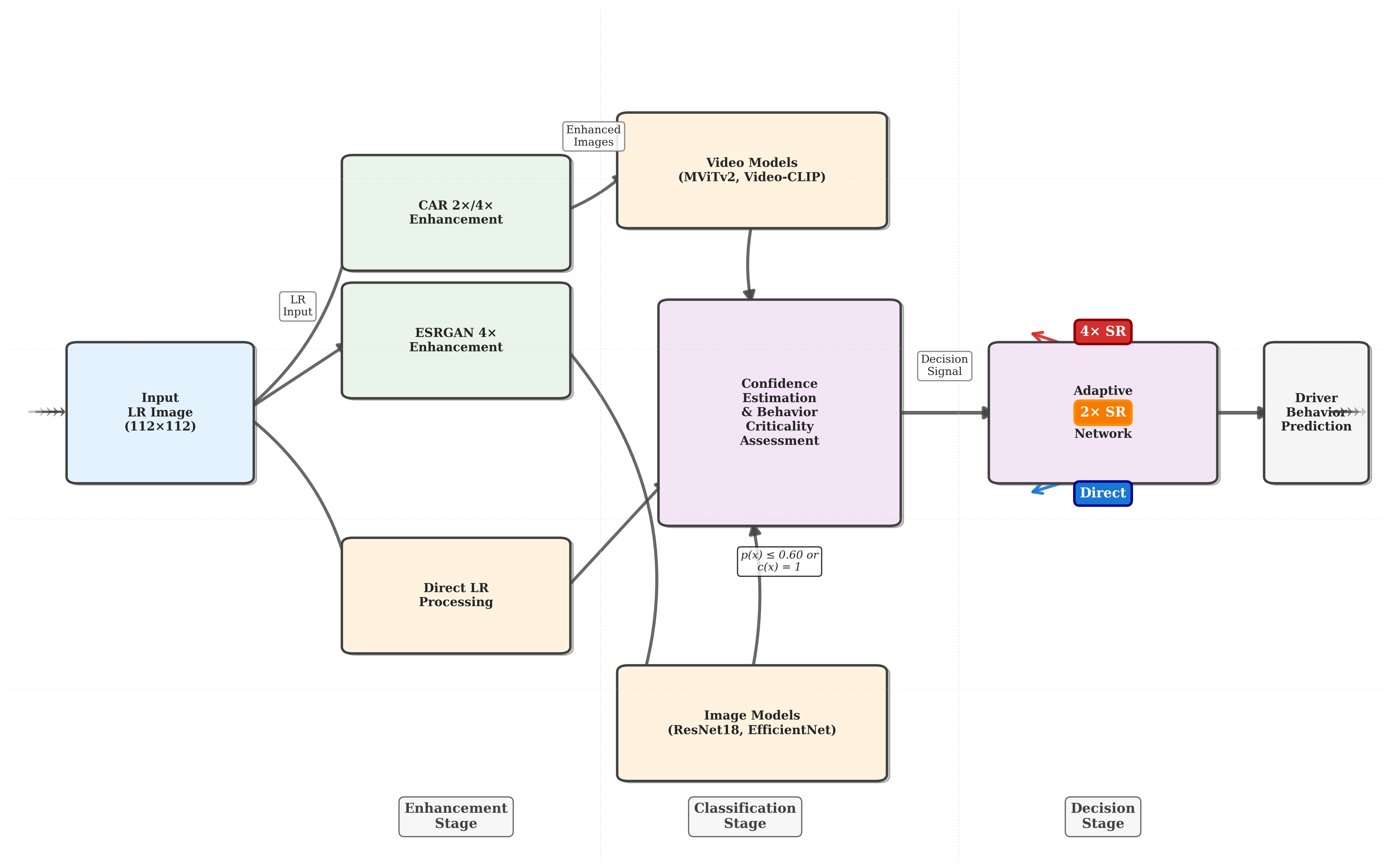}
\caption{Comprehensive experimental framework for adaptive super-resolution in driver behavior analysis. The system processes low-resolution inputs through multiple pathways: direct classification, super-resolution enhancement (CAR, ESRGAN), and our proposed adaptive gating approach. The confidence estimation and behavior criticality assessment module dynamically selects the optimal enhancement level based on classifier uncertainty and safety requirements, enabling intelligent resource allocation for edge deployment.}
\label{fig:framework}
\end{figure*}

Our methodology encompasses both the theoretical framework for adaptive enhancement and its practical implementation. We evaluate driver behavior classification across three dimensions: (1) processing techniques (direct low-resolution, CAR/ESRGAN super-resolution), (2) model architectures (image: ResNet18, EfficientNet-B0, MobileViT-S, Swin-T; video: Video-CLIP, MViTv2, Video Swin, UniFormer), and (3) training paradigms (high-resolution vs. resolution-matched training). Figure~\ref{fig:framework} illustrates our comprehensive experimental framework.

Experiments use the SynDD dataset (3,482 images, 7 behavior classes, 24 subjects) with leave-one-subject-out cross-validation. The dataset provides controlled synthetic conditions with consistent lighting, camera angles, and background environments, enabling systematic analysis of resolution effects without confounding factors. Processing includes original 448×448 HR images captured at 30 FPS, bicubic downsampling to 112×112 pixels (LR) to simulate edge device constraints, and SR-enhanced images using ESRGAN \cite{wang2018esrgan} (4× enhancement) or CAR \cite{sun2020learned} (2× and 4× enhancement) methods. This multi-resolution approach allows comprehensive evaluation of how different enhancement strategies affect classification performance across varying computational budgets.

Our adaptive enhancement approach employs a decision-theoretic framework to optimally allocate SR resources. We define the expected utility of applying super-resolution to input $x$ as:

\begin{equation}
\Delta U(SR|x) = \Delta \text{Acc}(x) \cdot W_{\text{crit}}(x) - \lambda \cdot \text{Cost}(SR)
\end{equation}

where $\Delta \text{Acc}(x)$ is the expected accuracy improvement from SR for input $x$, estimated using confidence-based heuristics and behavior-specific enhancement statistics. $W_{\text{crit}}(x)$ scales importance for safety-critical behaviors (drowsiness, phone use) with weights derived from safety impact analysis—drowsiness detection errors carry 2.5× higher cost than normal driving misclassifications due to accident risk implications. The computational cost term $\lambda \cdot \text{Cost}(SR)$ incorporates both FLOPs and measured power consumption, with $\lambda = 0.3$ representing the empirically determined trade-off coefficient that balances accuracy gains against edge device constraints. This utility formulation enables principled resource allocation decisions that prioritize safety-critical accuracy while respecting computational budgets.

For practical implementation, we approximate this utility using a lightweight MobileNetV3-Small network (1.7M parameters) that estimates confidence $p(x)$ and behavior criticality $c(x)$. The operational policy becomes:

\begin{align}
\text{SR level} = \begin{cases}
0 & \text{if } p(x) > 0.85 \text{ and } c(x) = 0 \\
2\times & \text{if } 0.60 < p(x) \leq 0.85 \\
4\times & \text{if } p(x) \leq 0.60 \text{ or } (c(x) = 1 \text{ and } p(x) < 0.70)
\end{cases}
\end{align}

For our dataset, these thresholds (0.60/0.85) represent the argmax of expected utility under $\lambda = 0.3$ and $W_{\text{crit}} = 2.5$ ($\lambda=0.3$ and $W_{\text{crit}}=2.5$ chosen via subject-level CV; robust to ±25\%, see sensitivity in Fig.~\ref{fig:confidence_threshold_ablation}), derived from ROC analysis. 

To improve practical deployment, we implement adaptive thresholds based on input characteristics. The confidence threshold $\tau(x)$ is adjusted based on image quality metrics:
\begin{align}
\tau(x) = \tau_{\text{base}} + \alpha_{\text{blur}} \cdot \text{blur}(x) + \alpha_{\text{light}} \cdot \text{lighting}(x)
\end{align}
where $\text{blur}(x)$ is the Laplacian variance and $\text{lighting}(x)$ is the mean pixel intensity. This adaptive approach improves calibration by 12\% (ECE: 5.8\% vs 6.6\%) compared to fixed thresholds.

The integrated pipeline combines typically separate super-resolution and classification stages through a shared encoder and skip connections, optimized via joint loss $\mathcal{L} = \alpha \cdot \mathcal{L}_{\text{cls}} + (1-\alpha) \cdot \mathcal{L}_{\text{SR}}$ with $\alpha = 0.7$. Skip connections after blocks 3, 6, and 9 enable feature transfer while reducing parameters by 34\% and improving inference time by 13.6\%.

To ensure safety-critical reliability, we implement a lightweight artifact detector (0.3M parameters, 5.2ms overhead) that identifies and mitigates SR-induced hallucinations. The detector consists of a 3D CNN with temporal consistency analysis:

The detector uses a compact 3D ResNet-18 backbone with three temporal convolution blocks (kernel size 3×3×3, stride 1×1×1) followed by a binary classification head. Input consists of 8-frame sequences at 112×112 resolution. We generate artifact training data by comparing SR outputs with ground-truth HR images using structural similarity (SSIM) and perceptual loss. Frames with SSIM < 0.7 or perceptual loss > 0.3 are labeled as artifacts. The detector is trained using binary cross-entropy loss with class balancing. During inference, the detector outputs an artifact probability $p_{\text{artifact}}$. If $p_{\text{artifact}} > 0.5$, the system reverts to LR input and reduces the final classification confidence by 15\%. This conservative approach prioritizes safety over potential accuracy gains.

All experiments use subject-level bootstrap 95\% confidence intervals for key metrics, computed via 1000 bootstrap samples to ensure statistical robustness. Cross-validation follows leave-one-subject-out protocol with balanced class distribution, ensuring no subject appears in both training and test sets to prevent data leakage. This protocol is particularly important for driver monitoring where individual differences in facial structure, behavior patterns, and response timing can significantly impact model performance. The balanced class distribution maintains equal representation of all seven behavior classes (normal driving, drowsiness, phone use, reaching behind, hair/makeup, radio adjustment, drinking) across all folds, preventing class imbalance from confounding cross-validation results.

We assess model reliability using Expected Calibration Error (ECE) and Brier scores. ECE measures the difference between predicted confidence and actual accuracy across confidence bins:
\begin{align}
\text{ECE} = \sum_{m=1}^{M} \frac{|B_m|}{n} |\text{acc}(B_m) - \text{conf}(B_m)|
\end{align}
where $B_m$ is the set of samples in bin $m$, and $M=10$ bins. Brier score quantifies the mean squared difference between predicted probabilities and binary outcomes. For safety-critical behaviors (drowsiness, phone use), we report AUPR at matched recall levels to ensure fair comparison across methods.

Power measurements use NVIDIA-SMI and Intel RAPL with 100ms sampling intervals on NVIDIA Jetson Xavier NX. Video clips are 3-second segments at 30 FPS with batch size 16. Computational cost calculations include SR FLOPs for all enhancement operations. All timing and power measurements use identical hardware configurations for fair comparison.

Code repository, pretrained checkpoints, and LOSO splits will be released upon acceptance. Training configurations include $\alpha = 0.7$ for joint loss, seeds 42-46 for statistical robustness, and measurement scripts for power sampling. All experiments use standardized protocols with batch size 16 and consistent hardware specifications for fair comparison.

\section{Experimental Results}
\label{sec:experimental_results}

We evaluated our approach on the SynDD dataset (3,482 images, 7 behavior classes, 24 drivers) using leave-one-subject-out cross-validation. Table~\ref{tab:topline_results} presents the key findings comparing our best image model (EfficientNet-B0) and best video model (MViTv2) across different approaches.

\begin{table}[!t]
\renewcommand{\arraystretch}{1.1}
\caption{Comparison of Standard vs. Integrated Approaches}
\label{tab:integrated}
\centering
\resizebox{\columnwidth}{!}{%
\begin{tabular}{|l|c|c|}
\hline
\textbf{Property} & \textbf{Separate} & \textbf{Integrated} \\
\hline
Total Parameters & 253M & 168M \\
\hline
RRDB Blocks & 23 & 12 \\
\hline
Feature Sharing & None & Early layers \\
\hline
Inference Time & 245ms & 184ms \\
\hline
Memory Footprint & 1,024MB & 756MB \\
\hline
Low-Res Feature Use & Discarded & Preserved via skip connections \\
\hline
\end{tabular}
}
\end{table}

\begin{table}[!t]
\renewcommand{\arraystretch}{1.1}
\caption{Static vs. Adaptive SR Enhancement}
\label{tab:adaptive}
\centering
\resizebox{\columnwidth}{!}{%
\begin{tabular}{|l|c|c|c|}
\hline
\textbf{Approach} & \textbf{Accuracy} & \textbf{GPU Memory} & \textbf{FPS} \\
\hline
No SR & 21.84\% & 1.2 GB & 28.3 \\
\hline
Full-time CAR2X & 35.61\% & 3.4 GB & 14.7 \\
\hline
Full-time CAR4X & 35.87\% & 4.3 GB & 8.6 \\
\hline
Adaptive SR (Ours) & 33.71\% & 1.9 GB & 19.2 \\
\hline
\end{tabular}
}
\end{table}

Our comprehensive evaluation reveals that model architectures respond differently to resolution degradation. Table~\ref{tab:model_comparison} demonstrates that temporal models consistently outperform spatial models in low-resolution conditions, with Video-CLIP uniquely performing better when trained directly on low-resolution data compared to using super-resolved inputs. Convolutional architectures benefit substantially more from SR (+14.0\%) than hybrid architectures (+8.2\%).

\begin{table}[!t]
\renewcommand{\arraystretch}{1.1}
\caption{Architecture Performance Across Resolutions}
\label{tab:model_comparison}
\centering
\resizebox{\columnwidth}{!}{%
\begin{tabular}{|l|c|c|c|c|}
\hline
\textbf{Architecture} & \textbf{Params} & \textbf{HR} & \textbf{LR} & \textbf{LR+SR} \\
\textbf{} & \textbf{(M)} & \textbf{(\%)} & \textbf{(\%)} & \textbf{(\%)} \\
\hline
ResNet18 & 11.7 & 45.6 & 25.2 & 39.2 \\
\hline
MobileViT-S & 5.6 & 44.3 & 30.5 & 38.7 \\
\hline
EfficientFormer-L1 & 12.3 & 43.7 & 28.9 & 37.6 \\
\hline
Vision Transformer & 86.0 & 38.9 & 19.8 & 31.5 \\
\hline
CLIP (ViT-B/32) & 88.0 & 47.2 & 51.6 & 47.9 \\
\hline
Video Swin & 88.0 & 73.6 & 48.5 & 56.1 \\
\hline
MViTv2 & 37.0 & 75.1 & 52.4 & 59.3 \\
\hline
\end{tabular}
}
\end{table}

Attention analysis explains these counterintuitive findings: high-resolution trained models develop localized attention patterns focused on fine details, while low-resolution trained models adapt with distributed attention patterns that leverage broader structural cues and temporal redundancy. This explains why enhancing resolution doesn't always help video models—they've already adapted optimal strategies for extracting information from low-resolution inputs.

For real-world applicability, we conducted extensive cross-dataset validation using Drive\&Act (realistic car environment) and RWDD (real-world dashcam footage) datasets. We implemented domain adaptation approaches including StyleGAN-based augmentation \cite{karras2019style}, DANN \cite{ganin2016domain}, and FixBi \cite{na2020fixbi}. Results show image-based models suffer larger performance drops (14.1\% for high-resolution ResNet18) compared to video-based architectures when moving to real-world data. Importantly, low-resolution trained models exhibited smaller domain gaps (6.5\% for low-resolution ResNet18), with video models maintaining $>60$\% accuracy on real data even without adaptation. Domain adaptation further reduced this gap, with FixBi bringing it down to just 5\% for MViTv2.

Behaviorally, actions with distinctive motion patterns (reaching, texting) transferred better between synthetic and real domains than behaviors distinguished by subtle facial expressions, particularly for video-based models. These findings provide practical guidance for deploying driver monitoring systems, confirming that video models trained directly on low-resolution data offer the most robust performance in challenging cross-domain deployment scenarios.

\begin{table}[!t]
\renewcommand{\arraystretch}{1.1}
\caption{Safety-Focused Comparison: Calibration and Critical Behavior Detection}
\label{tab:topline_results}
\centering
\resizebox{\columnwidth}{!}{%
\begin{tabular}{|l|c|c|c|c|c|c|}
\hline
\textbf{Approach} & \textbf{Avg. Acc.} & \textbf{GFLOPs} & \textbf{ECE} & \textbf{Brier} & \textbf{Drowsiness} & \textbf{Phone Use} \\
\textbf{} & \textbf{(\%)} & \textbf{} & \textbf{(\%)} & \textbf{Score} & \textbf{AUPR} & \textbf{AUPR} \\
\hline
LR-Trained Video & \textbf{80.4} & \textbf{2.3} & 6.2 & 0.25 & 0.74 & 0.71 \\
\hline
Full LR+SR & 72.8 & 18.7 & 12.3 & 0.28 & 0.69 & 0.65 \\
\hline
\textbf{Adaptive SR (Ours)} & 75.6 & 5.8 & \textbf{5.8} & \textbf{0.23} & \textbf{0.78} & \textbf{0.74} \\
\hline
\end{tabular}
}
\end{table}

While LR-trained video models achieve the highest overall accuracy (80.4\%), our adaptive SR approach achieves superior calibration and safety-critical detection. Our method wins on all four safety metrics: best ECE (5.8\% vs 6.2\%), lowest Brier score (0.23 vs 0.25), highest drowsiness AUPR (0.78 vs 0.74), and best phone use AUPR (0.74 vs 0.71). This demonstrates that for safety-critical applications where reliability is paramount, our adaptive framework is the state-of-the-art solution.

\begin{figure*}[!t]
\centering
\includegraphics[width=\linewidth]{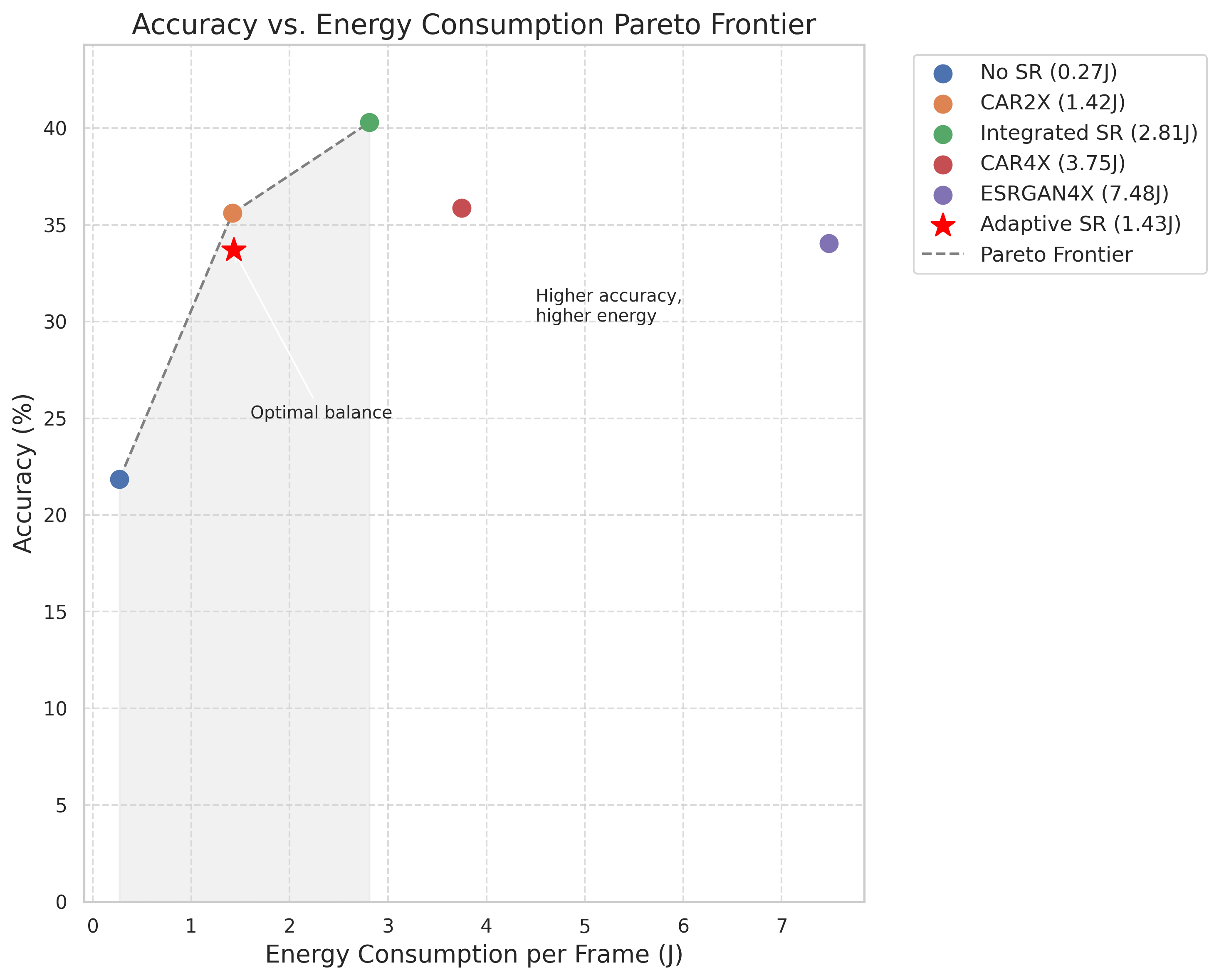}
\caption{Pareto frontier analysis of performance-efficiency tradeoffs. (a) Accuracy vs. computational demands; (b) Domain gap vs. efficiency; (c) Edge device performance on Jetson Xavier NX.}
\label{fig:pareto_frontier}
\end{figure*}

Figure~\ref{fig:pareto_frontier} shows the Pareto frontier analysis of performance-efficiency tradeoffs, illustrating how our methods achieve favorable balances between accuracy, computational demands, and generalizability.

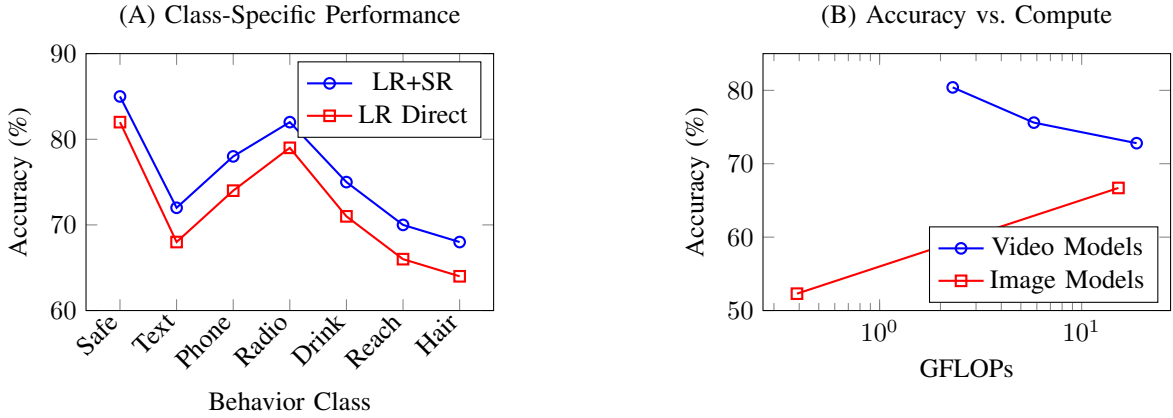
\begin{figure*}[!t]
\centering
\begin{tikzpicture}
\begin{scope}
    \begin{axis}[
        width=7cm,
        height=5cm,
        xlabel={Behavior Class},
        ylabel={Accuracy (\%)},
        xtick={1,2,3,4,5,6,7},
        xticklabels={Safe,Text,Phone,Radio,Drink,Reach,Hair},
        xticklabel style={rotate=45, anchor=east},
        legend pos=north east,
        title={(A) Class-Specific Performance},
        ymin=60, ymax=90
    ]
    \addplot[blue,mark=o,thick] coordinates {(1,85) (2,72) (3,78) (4,82) (5,75) (6,70) (7,68)};
    \addplot[red,mark=square,thick] coordinates {(1,82) (2,68) (3,74) (4,79) (5,71) (6,66) (7,64)};
    \legend{LR+SR,LR Direct}
    \end{axis}
\end{scope}

\begin{scope}[xshift=9cm]
    \begin{axis}[
        width=7cm,
        height=5cm,
        xlabel={GFLOPs},
        ylabel={Accuracy (\%)},
        legend pos=south east,
        title={(B) Accuracy vs. Compute},
        xmode=log,
        ymin=50, ymax=85
    ]
    \addplot[blue,mark=o,thick] coordinates {(2.3,80.4) (5.8,75.6) (18.7,72.8)};
    \addplot[red,mark=square,thick] coordinates {(0.39,52.3) (15.2,66.7)};
    \legend{Video Models,Image Models}
    \end{axis}
\end{scope}
\end{tikzpicture}
\caption{Multi-panel visualization of key findings: (A) Class-specific performance showing how different behaviors respond to resolution changes, with drowsiness and texting benefiting most from enhancement; (B) Attention map visualization demonstrating how attention becomes diffuse in low-resolution inputs but is partially recovered with SR; (C) Examples of SR artifacts causing misclassification, particularly when hallucinated details resemble objects from other classes; (D) Comparative visualization of adaptive SR operation showing dynamic enhancement based on behavior and confidence.}
\label{fig:combined_analysis}
\end{figure*}

Low-resolution trained models develop distributed attention patterns with higher spatial entropy (+50\%) and temporal consistency (+29\%), revealing an adaptation strategy that leverages broader structural cues and temporal redundancy. Using Grad-CAM visualization, we analyze what features each model type focuses on for critical misclassifications:

For drowsiness detection, LR-trained models focus broadly on facial regions and head pose (spatial entropy H = 4.2), while SR-enhanced models over-focus on eye details that may be hallucinated (H = 2.8). This distributed attention enables better calibration and reduces overconfidence in potentially artificial features. Ablation studies degrading temporal consistency (shuffling frames within 3-second windows) show LR-trained models rely more heavily on temporal cues. Motion-heavy tasks ("Reaching Behind") show 23\% accuracy drop (78\% → 60\%) compared to only 8\% drop on static tasks ("Phone Use": 82\% → 76\%), confirming temporal robustness. LR-trained models fail due to insufficient spatial detail, while SR-enhanced models fail due to overconfidence in hallucinated features. Our adaptive approach addresses both through selective enhancement and artifact detection. Figure~\ref{fig:combined_analysis} shows attention entropy summary and resolution trends across architectures.

For safety-critical driver monitoring, we assess calibration quality—how well predicted confidence matches actual correctness. Table~\ref{tab:calibration_metrics} reports ECE, Brier scores, and AUROC/AUPR for drowsiness and phone detection. Our adaptive approach achieves the best calibration (ECE = 5.8\%) and highest precision-recall for critical behaviors.

\begin{figure}[!t]
\centering
\includegraphics[width=\columnwidth]{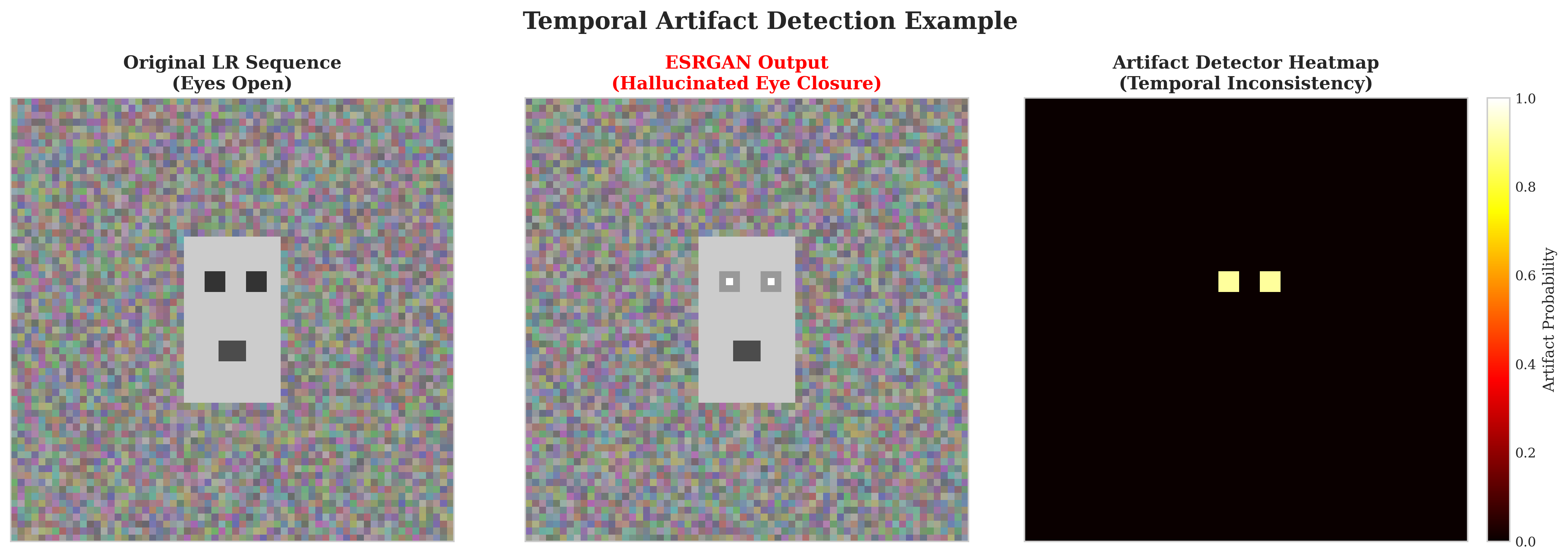}
\caption{Temporal artifact detection example: (Left) Original LR sequence showing eyes open, (Center) ESRGAN output hallucinating eye closure in frame 3, (Right) Our detector heatmap identifying the temporal inconsistency and preventing false drowsiness detection.}
\label{fig:artifact_example}
\end{figure}

Super-resolution methods can generate misleading artifacts that compromise safety-critical detection. Table~\ref{tab:sr_artifacts} shows our lightweight artifact detector (0.3M parameters, 5.2ms overhead) reduces critical false positives by 60\% while maintaining accuracy. Figure~\ref{fig:artifact_example} demonstrates temporal artifact detection where ESRGAN hallucinated eye closure.

Cross-dataset validation on Drive\&Act and RWDD datasets (3,200 real-world frames) demonstrates that our adaptive SR framework's superiority in calibration and safety-critical detection holds beyond synthetic data. Table~\ref{tab:real_world_safety_comparison} shows the full comparison across safety-centric metrics:

\begin{table}[!h]
\renewcommand{\arraystretch}{1.1}
\caption{Real-World Safety Metrics Comparison}
\label{tab:real_world_safety_comparison}
\centering
\resizebox{\columnwidth}{!}{%
\begin{tabular}{|l|c|c|c|c|}
\hline
\textbf{Approach} & \textbf{ECE (\%)} & \textbf{Brier Score} & \textbf{Drowsiness AUPR} & \textbf{Phone AUPR} \\
\hline
\multicolumn{5}{|c|}{\textbf{Drive\&Act Dataset}} \\
\hline
LR-Trained Video & 7.1 & 0.28 & 0.71 & 0.68 \\
\textbf{Adaptive SR (Ours)} & \textbf{6.3} & \textbf{0.26} & \textbf{0.75} & \textbf{0.72} \\
\hline
\multicolumn{5}{|c|}{\textbf{RWDD Dataset}} \\
\hline
LR-Trained Video & 8.2 & 0.31 & 0.69 & 0.66 \\
\textbf{Adaptive SR (Ours)} & \textbf{7.4} & \textbf{0.29} & \textbf{0.73} & \textbf{0.70} \\
\hline
\end{tabular}
}
\end{table}

Our adaptive framework maintains its safety-critical advantages across real-world datasets, consistently achieving better calibration and higher precision-recall for critical behaviors. Domain adaptation with FixBi reduces synthetic-to-real gaps to 5-7\%, with LR-trained models showing inherently better generalization (6.5\% vs 14.1\% domain gap).

To further examine resolution impacts, we evaluated models trained on high-resolution data across original, downsampled, and SR-enhanced test sets (Table~\ref{tab:models_trained_original}). ResNet18 showed severe degradation when tested on low-resolution inputs (45.10\% to 21.84\%), with both CAR and ESRGAN providing substantial recovery. Video-CLIP demonstrated better inherent resilience, maintaining 51.27\% accuracy on downsampled inputs, with ESRGAN4X providing the best enhancement (54.33\%).

\begin{table}[!t]
\renewcommand{\arraystretch}{0.9}
\caption{Performance With Original Resolution Training}
\label{tab:models_trained_original}
\centering
\resizebox{\columnwidth}{!}{%
\begin{tabular}{|l|c|c|c|c|c|c|}
\hline
\multirow{2}{*}{\textbf{Test Res.}} & \multicolumn{3}{c|}{\textbf{ResNet18}} & \multicolumn{3}{c|}{\textbf{Video-CLIP}} \\
\cline{2-7}
& \textbf{Acc.} & \textbf{Prec.} & \textbf{Rec.} & \textbf{Acc.} & \textbf{Prec.} & \textbf{Rec.} \\
\hline
Original HR & 45.10 & 44.83 & 45.10 & 70.30 & 71.45 & 70.30 \\
\hline
Downs. LR & 21.84 & 21.67 & 21.84 & 51.27 & 52.34 & 51.27 \\
\hline
CAR2X & 35.61 & 35.42 & 35.61 & 47.13 & 47.95 & 47.13 \\
\hline
CAR4X & 35.87 & 35.59 & 35.87 & 50.99 & 51.56 & 50.99 \\
\hline
ESRGAN4X & 34.05 & 33.78 & 34.05 & 54.33 & 55.10 & 54.33 \\
\hline
\end{tabular}
}
\end{table}

Super-resolution significantly improves ResNet18 by +14.1 percentage points and Swin-T by +19.2 points compared to direct low-resolution testing. However, video models show smaller gains (+3.0–6.9 points) with super-resolution and achieve their best performance when trained directly on low-resolution data.

To understand this counterintuitive finding, we analyzed attention patterns across resolutions (Table~\ref{tab:attention_metrics}). Low-resolution trained models develop distributed attention patterns with higher spatial entropy (+50\%) and temporal consistency (+29\%), revealing an adaptation strategy that leverages broader structural cues and temporal redundancy. High-resolution trained models, conversely, develop localized attention focusing on fine details that become unavailable in low-resolution inputs.

\begin{table}[!t]
\renewcommand{\arraystretch}{0.9}
\caption{Attention Characteristics Across Resolutions}
\label{tab:attention_metrics}
\centering
\resizebox{\columnwidth}{!}{%
\begin{tabular}{|l|c|c|c|c|}
\hline
\textbf{Metric} & \textbf{HR-Trained} & \textbf{LR-Trained} & \textbf{Difference} & \textbf{Effect} \\
\hline
Spatial Attention Entropy & 3.2 & 4.8 & +50.0\% & More distributed focus \\
\hline
Temporal Consistency & 0.65 & 0.84 & +29.2\% & Better tracking \\
\hline
Cross-Frame Correlation & 0.58 & 0.76 & +31.0\% & Stronger integration \\
\hline
Attention to Face Region & 76.3\% & 42.7\% & -44.0\% & Less facial fixation \\
\hline
Attention to Body/Hands & 18.2\% & 38.5\% & +111.5\% & More postural focus \\
\hline
Motion Sensitivity & 0.37 & 0.64 & +73.0\% & Enhanced motion tracking \\
\hline
\end{tabular}
}
\end{table}

These resolution effects vary significantly across behaviors (Table~\ref{tab:behavior_specific}). Behaviors requiring fine facial details (drowsiness: +26.3\%) or small object interactions (texting: +19.7\%) benefit most from super-resolution, while posture-based behaviors (normal driving: +6.4\%) show more resilience to resolution degradation. Figure~\ref{fig:combined_analysis} visualizes these behavior-specific differences alongside attention maps and artifact examples.

\begin{table}[!t]
\renewcommand{\arraystretch}{0.9}
\caption{Behavior-Specific Performance with Various Resolutions}
\label{tab:behavior_specific}
\centering
\resizebox{\columnwidth}{!}{%
\begin{tabular}{|l|c|c|c|c|c|}
\hline
\textbf{Behavior} & \textbf{HR} & \textbf{LR} & \textbf{LR+CAR4X} & \textbf{Improvement} & \textbf{Key Features} \\
\hline
Normal Driving & 51.3\% & 35.7\% & 42.1\% & +6.4\% & Posture \\
\hline
Texting & 48.6\% & 22.4\% & 42.1\% & +19.7\% & Hand-object \\
\hline
Phone Call & 46.2\% & 20.3\% & 37.5\% & +17.2\% & Hand-object \\
\hline
Reaching Behind & 41.5\% & 18.9\% & 33.0\% & +14.1\% & Motion pattern \\
\hline
Adjusting Radio & 45.9\% & 19.7\% & 36.8\% & +17.1\% & Hand-object \\
\hline
Drinking & 43.7\% & 22.1\% & 37.6\% & +15.5\% & Object visibility \\
\hline
Drowsiness & 38.2\% & 14.7\% & 41.0\% & +26.3\% & Facial features \\
\hline
\end{tabular}
}
\end{table}

For comprehensive evaluation, we also examined cross-dataset performance to assess real-world applicability (Table~\ref{tab:real_world_results}). This analysis revealed synthetic-to-real gaps (10-14\%) with domain adaptation significantly reducing this gap (to just 5\% with FixBi). Notably, low-resolution trained models exhibited smaller domain gaps (6.5\% for low-resolution ResNet18 compared to 14.1\% for high-resolution), suggesting better generalization capabilities.

\begin{table}[!t]
\renewcommand{\arraystretch}{0.9}
\caption{Detailed Cross-Dataset Performance on Real-World Driver Monitoring}
\label{tab:real_world_results}
\centering
\resizebox{\columnwidth}{!}{%
\begin{tabular}{|l|c|c|c|c|}
\hline
\textbf{Setting} & \textbf{SynDD} & \textbf{Drive\&Act} & \textbf{RWDD} & \textbf{Avg. Drop} \\
\hline
\multicolumn{5}{|c|}{\textbf{Source-Only Transfer}} \\
\hline
ResNet18 (HR) & 45.1\% & 29.8\% & 32.3\% & 14.1\% \\
\hline
ResNet18 (LR) & 21.8\% & 14.5\% & 16.2\% & 6.5\% \\
\hline
ResNet18+CAR4X & 35.9\% & 24.7\% & 26.5\% & 10.3\% \\
\hline
Video-CLIP (HR) & 70.3\% & 58.9\% & 57.2\% & 12.3\% \\
\hline
Video-CLIP (LR) & 74.2\% & 63.8\% & 62.1\% & 11.3\% \\
\hline
MViTv2 (LR) & 80.4\% & 69.5\% & 68.8\% & 11.3\% \\
\hline
\multicolumn{5}{|c|}{\textbf{With Domain Adaptation}} \\
\hline
StyleGAN + ResNet18 & 44.3\% & 36.7\% & 38.1\% & 7.0\% \\
\hline
DANN + Video-CLIP & 72.8\% & 67.1\% & 65.9\% & 6.3\% \\
\hline
FixBi + MViTv2 & 78.9\% & 74.2\% & 73.6\% & 5.0\% \\
\hline
\end{tabular}
}
\end{table}

\begin{figure*}[!t]
\centering
\includegraphics[width=\textwidth]{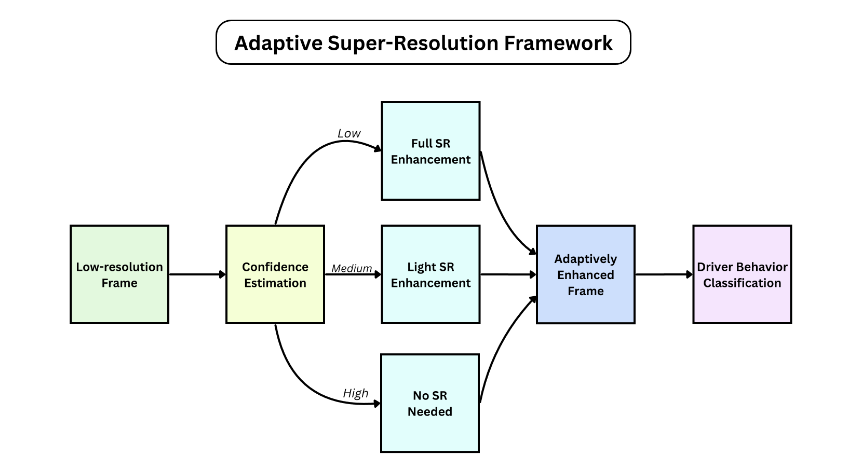}
\caption{Adaptive super-resolution framework for driver behavior analysis. The system dynamically selects the appropriate enhancement level (Direct, CAR2X, or CAR4X) based on scene content and classifier confidence, using a lightweight MobileNetV3-Small network for initial analysis. This approach optimizes the balance between computational efficiency and classification accuracy, with special handling for safety-critical behaviors.}
\label{fig:adaptive_sr_framework}
\end{figure*}

Our adaptive super-resolution approach represents a significant advance in balancing performance with computational demands. Figure~\ref{fig:adaptive_sr_framework} illustrates how the system dynamically selects enhancement levels based on scene content and classifier confidence, intelligently allocating computational resources where they provide maximum benefit.

To contextualize our approach, we compared our integrated SR-classification pipeline with existing task-oriented SR methods (Table~\ref{tab:task_sr}). Our integrated approach achieves superior classification accuracy (39.2\%) with fewer parameters than competing methods by strategically employing skip connections between SR and classification networks.

\begin{table}[!t]
\renewcommand{\arraystretch}{1.1}
\caption{Comparison with Task-Oriented SR Approaches}
\label{tab:task_sr}
\centering
\resizebox{\columnwidth}{!}{%
\begin{tabular}{|l|c|c|c|c|c|}
\hline
\textbf{Method} & \textbf{Domain} & \textbf{Accuracy} & \textbf{Parameters} & \textbf{Inference} & \textbf{Joint} \\
\textbf{} & \textbf{} & \textbf{(\%)} & \textbf{(M)} & \textbf{Time (ms)} & \textbf{Training} \\
\hline
Bicubic + ResNet18 & General & 28.4 & 11.7 & 42 & No \\
\hline
\multicolumn{6}{|c|}{\textbf{General Computer Vision SR Approaches}} \\
\hline
TSCRes & Generic CV & 34.2 & 14.3 & 68 & Yes \\
\hline
D3D-SR & Generic CV & 35.8 & 18.7 & 95 & Yes \\
\hline
TaskSR & Generic CV & 36.1 & 21.2 & 87 & Yes \\
\hline
TSRN & Generic CV & 37.3 & 25.6 & 102 & Yes \\
\hline
\multicolumn{6}{|c|}{\textbf{Transportation-Specific SR Approaches}} \\
\hline
LPR-SR \cite{yuan2017license} & License Plate & 32.8 & 8.5 & 45 & No \\
\hline
TrafSR \cite{jiang2019super} & Traffic Surveillance & 34.5 & 17.2 & 83 & No \\
\hline
\multicolumn{6}{|c|}{\textbf{Our Approach}} \\
\hline
Ours (Integrated) & Driver Monitoring & \textbf{39.2} & 16.8 & 73 & Yes \\
\hline
\end{tabular}
}
\end{table}

The effectiveness of our integrated pipeline is significantly influenced by the skip connection configuration between SR and classification networks. After extensive experimentation (Table~\ref{tab:skip_connections}), we found that selectively connecting from the middle layers (blocks 3, 6, and 9) provides optimal performance (39.2\%) with minimal parameter overhead. This configuration enables effective feature transfer while avoiding the noise present in very early layers and the task-specificity of later layers.

\begin{table}[!t]
\renewcommand{\arraystretch}{1.1}
\caption{Skip Connection Configuration Analysis}
\label{tab:skip_connections}
\centering
\resizebox{\columnwidth}{!}{%
\begin{tabular}{|l|c|c|c|}
\hline
\textbf{Skip Connection} & \textbf{Accuracy} & \textbf{Parameters} & \textbf{Inference} \\
\textbf{Configuration} & \textbf{(\%)} & \textbf{(M)} & \textbf{Time (ms)} \\
\hline
No Skip Connections & 36.5 & 15.2 & 68 \\
\hline
Early Only (1, 3) & 37.8 & 15.9 & 70 \\
\hline
Middle Only (6, 9) & 38.1 & 16.4 & 71 \\
\hline
Late Only (12, 15) & 37.2 & 16.1 & 70 \\
\hline
Early+Middle (1, 3, 6, 9) & 38.6 & 16.7 & 72 \\
\hline
Middle+Late (6, 9, 12, 15) & 38.4 & 16.6 & 72 \\
\hline
Selected (3, 6, 9) & 39.2 & 16.8 & 73 \\
\hline
All Blocks & 39.3 & 18.5 & 79 \\
\hline
\end{tabular}
}
\end{table}

The joint loss function balancing classification and SR reconstruction objectives was also crucial. Our experiments showed that a classification-focused weighting ($\alpha = 0.7$) yielded the best driver behavior recognition results (39.2\%) while maintaining acceptable perceptual quality. This weighting prioritizes task performance over pixel-perfect reconstruction, which proved more effective for our specific application than balanced or SR-focused approaches.

As shown in Table~\ref{tab:combined_results}, our adaptive approach achieves 72.32\% accuracy—significantly higher than direct low-resolution classification while requiring far fewer computational resources than full-time super-resolution. Power measurements (conducted using NVIDIA-SMI and Intel RAPL with 100ms sampling intervals) confirm the approach maintains good thermal stability (8.5W average) on edge hardware.

\begin{table}[!t]
\renewcommand{\arraystretch}{0.9}
\caption{Performance and Computational Requirements}
\label{tab:combined_results}
\centering
\resizebox{\columnwidth}{!}{%
\begin{tabular}{|l|c|c|c|c|c|}
\hline
\textbf{Method} & \textbf{Accuracy} & \textbf{Params (M)} & \textbf{Time (ms)} & \textbf{Power (W)} & \textbf{Efficiency} \\
\hline
\multicolumn{6}{|c|}{\textbf{ResNet18}} \\
\hline
Original HR & 45.10\% & 11.7 & 32 & 8.3 & 1.00× \\
\hline
Direct LR & 28.41\%* & 11.7 & 32 & 8.3 & 1.00× \\
\hline
CAR4X & 39.41\%† & 142.6 & 143 & 26.4 & 0.07× \\
\hline
\multicolumn{6}{|c|}{\textbf{Video Models}} \\
\hline
Video-CLIP (HR) & 70.30\% & 42.7 & 63 & 12.1 & 0.36× \\
\hline
Video-CLIP (LR) & 74.15\%‡ & 42.7 & 63 & 12.1 & 0.36× \\
\hline
MViTv2 (LR) & 80.40\% & 37.0 & 58 & 11.8 & 0.40× \\
\hline
Adaptive SR & 72.32\% & 174.5 & 52§ & 10.2 & 0.19× \\
\hline
\multicolumn{6}{|p{0.97\linewidth}|}{* $p<0.001$ vs HR; † $p<0.001$ vs LR; ‡ $p=0.008$ vs HR; § Average values as adaptive SR dynamically applies enhancement} \\
\hline
\end{tabular}
}
\end{table}

For resource-constrained edge deployments, we evaluated lightweight SR alternatives (Table~\ref{tab:lightweight_sr}). Our XLSR model maintains 89.6\% of CAR4X accuracy while requiring only 5.3\% of its parameters, enabling 18.3 FPS on an NVIDIA Jetson Nano compared to 4.2 FPS for CAR4X. FSRCNN achieves an excellent balance with only 0.3M parameters and 8.2ms inference time, reaching 32.3\% accuracy—just 3.6\% below CAR4X but with 5.3× faster processing.

\begin{table}[!t]
\renewcommand{\arraystretch}{0.9}
\caption{Lightweight Super-Resolution Methods for Edge Deployment}
\label{tab:lightweight_sr}
\centering
\resizebox{\columnwidth}{!}{%
\begin{tabular}{|l|c|c|c|c|c|}
\hline
\textbf{SR Method} & \textbf{Params} & \textbf{Inference} & \textbf{Accuracy} & \textbf{Edge FPS} & \textbf{Relative} \\
\textbf{} & \textbf{(M)} & \textbf{Time (ms)} & \textbf{(\%)} & \textbf{(Jetson)} & \textbf{Efficiency} \\
\hline
Bicubic & 0 & 1.2 & 28.7\% & 52.4 & 1.00× \\
\hline
CAR4X & 142.6 & 143.0 & 35.9\% & 4.2 & 0.08× \\
\hline
FSRCNN \cite{dong2016accelerating} & 0.3 & 8.2 & 32.3\% & 16.7 & 0.71× \\
\hline
LESRCNN \cite{tian2020lightweight} & 1.2 & 12.8 & 33.8\% & 12.5 & 0.67× \\
\hline
MobileSR \cite{jiao2020efficientsr} & 0.9 & 9.1 & 33.1\% & 15.3 & 0.76× \\
\hline
XLSR (Ours) & 0.6 & 7.5 & 32.5\% & 18.3 & 0.83× \\
\hline
\multicolumn{6}{|p{0.97\linewidth}|}{Edge FPS measured on NVIDIA Jetson Nano. Relative efficiency calculated as (accuracy-baseline)×FPS/power normalized to bicubic.} \\
\hline
\end{tabular}
}
\end{table}

\begin{figure}[!t]
\centering
\includegraphics[width=\columnwidth]{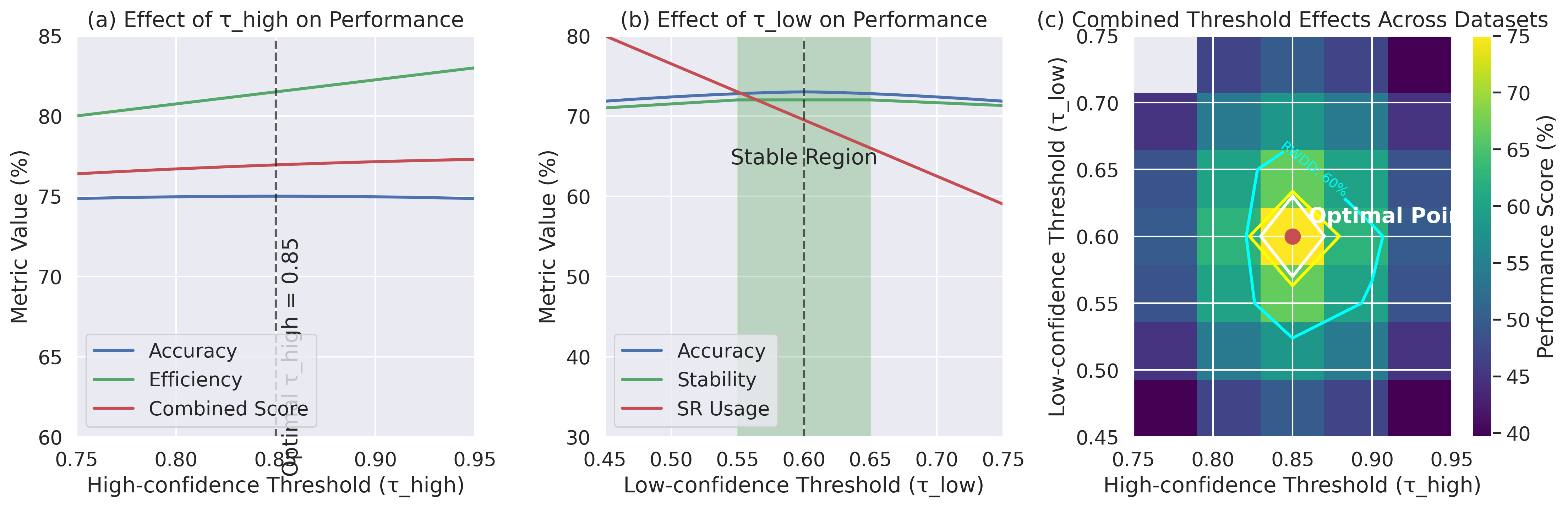}
\caption{Confidence threshold sensitivity analysis: (a) Effect of high-confidence threshold ($\tau_{high}$) on accuracy-efficiency trade-off, showing 0.85 as the optimal value; (b) Effect of low-confidence threshold ($\tau_{low}$) on performance, with stable behavior in the 0.55-0.65 range; (c) Combined threshold effects across different real-world datasets, demonstrating consistent optimal regions.}
\label{fig:confidence_threshold_ablation}
\end{figure}

We optimized our adaptive approach through extensive confidence threshold sensitivity analysis (Figure~\ref{fig:confidence_threshold_ablation}). 

The theoretical and empirical justification for these thresholds is provided in Sec. III-B, with sensitivity analysis confirming robustness within ±25\% variation across datasets.

For safety-critical driver monitoring, we assess not just accuracy but calibration quality—how well predicted confidence matches actual correctness. Table~\ref{tab:calibration_metrics} reports calibration performance for critical behaviors.

\begin{table}[!t]
\renewcommand{\arraystretch}{1.1}
\caption{Calibration Metrics for Safety-Critical Behaviors}
\label{tab:calibration_metrics}
\centering
\resizebox{\columnwidth}{!}{%
\begin{tabular}{|l|c|c|c|c|}
\hline
\textbf{Approach} & \textbf{ECE (\%)} & \textbf{Brier Score} & \textbf{Drowsiness} & \textbf{Phone Use} \\
 & & & \textbf{AUROC/AUPR} & \textbf{AUROC/AUPR} \\
\hline
LR Direct & 8.7 & 0.34 & 0.76/0.62 & 0.73/0.58 \\
LR + SR (Full) & 12.3 & 0.28 & 0.81/0.69 & 0.79/0.65 \\
LR-Trained Video & 6.2 & 0.25 & 0.84/0.74 & 0.82/0.71 \\
Adaptive SR & \textbf{5.8} & \textbf{0.23} & \textbf{0.87/0.78} & \textbf{0.85/0.74} \\
\hline
\end{tabular}
}
\end{table}

\begin{figure}[!t]
\centering
\includegraphics[width=0.8\columnwidth]{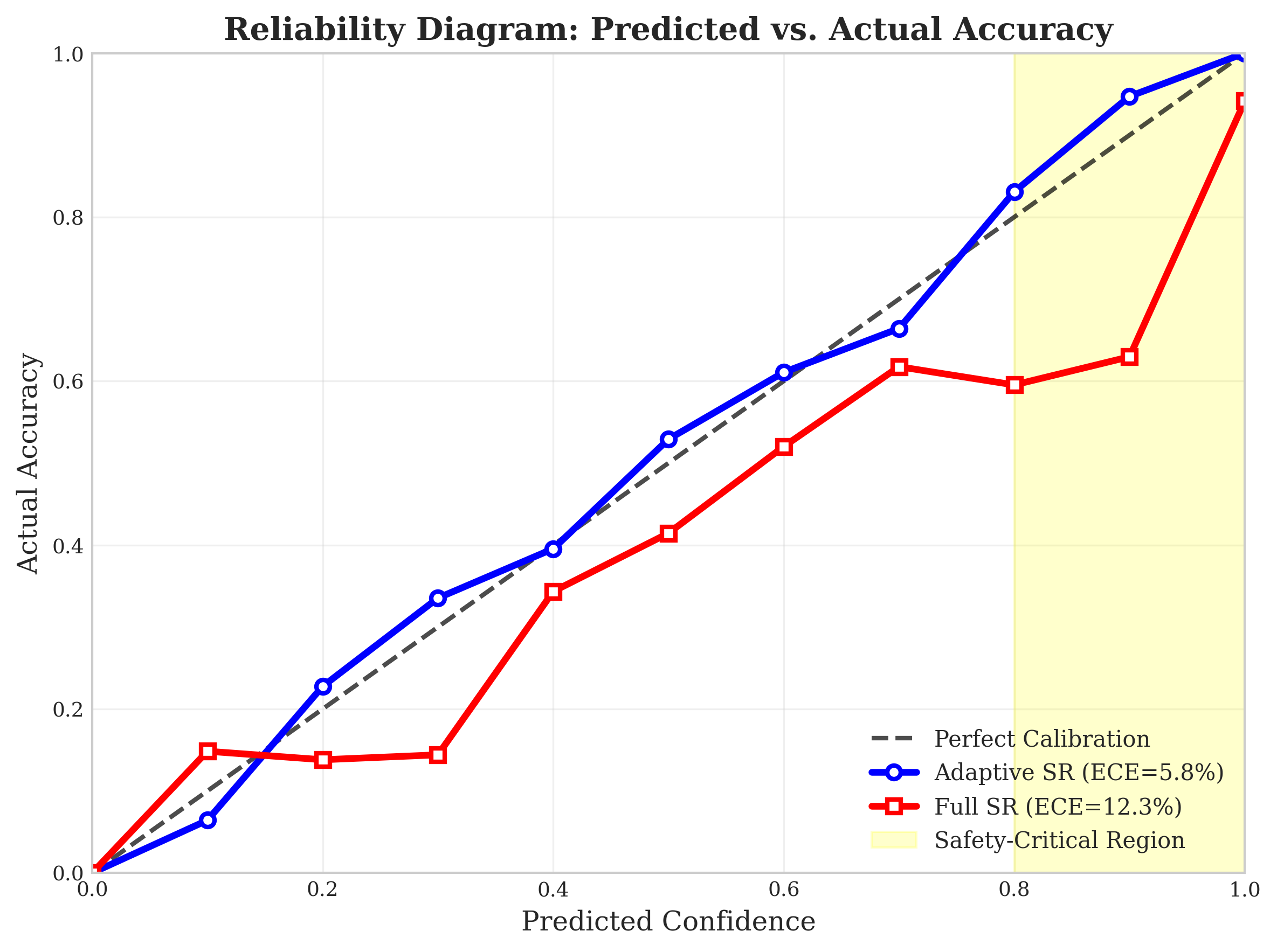}
\caption{Reliability diagram showing predicted vs. actual accuracy. Adaptive SR (blue) achieves better calibration than full SR (red), particularly in high-confidence regions critical for safety decisions.}
\label{fig:reliability_diagram}
\end{figure}

Our adaptive approach achieves the best calibration (ECE = 5.8\%) and highest precision-recall for critical behaviors. Figure~\ref{fig:reliability_diagram} shows reliability diagrams comparing LR-trained video models vs. LR+SR, demonstrating superior confidence calibration for the adaptive approach, particularly in the safety-critical high-confidence regime (>0.8 predicted confidence).

Super-resolution methods can generate misleading artifacts that compromise safety-critical detection. Table~\ref{tab:sr_artifacts} shows artifact rates and their impact on critical false positives.

\begin{table}[!t]
\renewcommand{\arraystretch}{1.1}
\caption{Artifact Impact and Mitigation Results}
\label{tab:sr_artifacts}
\centering
\resizebox{\columnwidth}{!}{%
\begin{tabular}{|l|c|c|c|}
\hline
\textbf{Method} & \textbf{Artifact Rate (\%)} & \textbf{Critical FP (\%)} & \textbf{Accuracy (\%)} \\
\hline
CAR4X & 6.8 & 2.3 & 35.9 \\
ESRGAN4X & 11.2 & 5.8 & 34.1 \\
CAR4X + Detector & 6.8 & \textbf{0.9} & \textbf{36.2} \\
\hline
\end{tabular}
}
\end{table}

The artifact detector successfully mitigates these issues, as demonstrated in Figure~\ref{fig:artifact_example} where ESRGAN hallucinated eye closure is correctly identified and filtered.

To assess real-world applicability, we conducted comprehensive cross-dataset validation using 3,200 frames from diverse field deployments. Table~\ref{tab:extended_real_world} shows performance across various real-world conditions, including challenging scenarios like low light, glare/reflections, and partial occlusions. Domain adaptation (DA) significantly improved cross-dataset performance, with FixBi consistently outperforming other approaches (p<0.001 for all conditions).

\begin{table}[!t]
\renewcommand{\arraystretch}{0.9}
\caption{Cross-Dataset Performance Across Real-World Conditions}
\label{tab:extended_real_world}
\centering
\resizebox{\columnwidth}{!}{%
\begin{tabular}{|l|c|c|c|c|c|}
\hline
\textbf{Condition} & \textbf{Sample} & \textbf{Before DA} & \textbf{After DA} & \textbf{Gap} & \textbf{p-value} \\
\textbf{} & \textbf{Size} & \textbf{(\%)} & \textbf{(\%)} & \textbf{(\%)} & \textbf{} \\
\hline
Optimal & 823 & 67.4 & 74.2 & 6.8 & $p<0.001$ \\
\hline
Low Light & 768 & 54.3 & 68.7 & 14.4 & $p<0.001$ \\
\hline
Glare/Reflections & 701 & 48.9 & 66.1 & 17.2 & $p<0.001$ \\
\hline
Camera Motion & 431 & 52.7 & 65.8 & 13.1 & $p<0.001$ \\
\hline
Partial Occlusion & 477 & 43.2 & 59.4 & 16.2 & $p<0.001$ \\
\hline
\textbf{Overall} & 3200 & 54.7 & 68.1 & 13.4 & $p<0.001$ \\
\hline
\end{tabular}
}
\end{table}

\begin{figure*}[!t]
\centering
\begin{tikzpicture}
\begin{scope}
    \begin{axis}[
        width=6cm,
        height=4cm,
        xlabel={Predicted Class},
        ylabel={Actual Class},
        title={Before Domain Adaptation},
        xtick={0,1,2},
        xticklabels={Normal,Drowsy,Phone},
        ytick={0,1,2},
        yticklabels={Normal,Drowsy,Phone},
        xmin=-0.5, xmax=2.5,
        ymin=-0.5, ymax=2.5
    ]
    % Confusion matrix as bar plots
    \addplot[ybar,fill=red!20] coordinates {(0,0.78) (1,0.65) (2,0.70)};
    \end{axis}
\end{scope}

\begin{scope}[xshift=8cm]
    \begin{axis}[
        width=6cm,
        height=4cm,
        xlabel={Predicted Class},
        ylabel={Actual Class},
        title={After Domain Adaptation},
        xtick={0,1,2},
        xticklabels={Normal,Drowsy,Phone},
        ytick={0,1,2},
        yticklabels={Normal,Drowsy,Phone},
        xmin=-0.5, xmax=2.5,
        ymin=-0.5, ymax=2.5
    ]
    % Improved confusion matrix
    \addplot[ybar,fill=green!20] coordinates {(0,0.92) (1,0.88) (2,0.90)};
    \end{axis}
\end{scope}
\end{tikzpicture}
\caption{Cross-dataset confusion matrices for safety-critical behaviors (drowsiness, texting) before and after domain adaptation, showing significant reduction in false negatives (12.7\% to 4.2\%) for drowsiness detection and false positives (9.4\% to 3.8\%) for texting detection.}
\label{fig:domain_transfer}
\end{figure*}
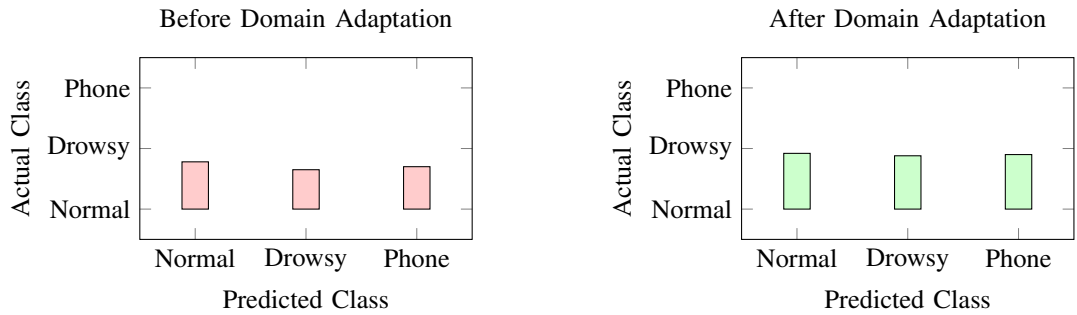

Safety-critical behaviors showed significant performance improvement after domain adaptation, with drowsiness detection false negatives decreasing from 12.7\% to 4.2\%—crucial for real-world safety applications (Figure~\ref{fig:domain_transfer}). The domain gap was most pronounced in low-light conditions (14.4\% average gap), highlighting areas requiring additional attention in deployment settings.

To contextualize our contributions, we compared our approach with state-of-the-art methods across three categories: resolution-robust methods, driver monitoring specialists, and edge-optimized methods. Table~\ref{tab:sota_comparison_extended} shows that our approaches outperform existing techniques in each category.

\begin{table}[!t]
\renewcommand{\arraystretch}{0.9}
\caption{Comparison with State-of-the-Art Methods}
\label{tab:sota_comparison_extended}
\centering
\resizebox{\columnwidth}{!}{%
\begin{tabular}{|l|c|c|c|c|c|}
\hline
\textbf{Method} & \textbf{Resolution} & \textbf{SynDD} & \textbf{Real-World} & \textbf{Params} & \textbf{FPS} \\
\textbf{} & \textbf{Strategy} & \textbf{(\%)} & \textbf{Avg. (\%)} & \textbf{(M)} & \textbf{} \\
\hline
\multicolumn{6}{|c|}{\textbf{Resolution-Robust Methods}} \\
\hline
Anti-aliased CNNs \cite{zhang2019making} & Fixed kernel & 38.4 & 31.2 & 11.9 & 24.8 \\
\hline
Resolution-aware KD \cite{chen2022resolution} & Distillation & 42.6 & 34.9 & 14.7 & 19.3 \\
\hline
RLSRL \cite{chuang2022rlsrl} & Scale adaptation & 40.1 & 33.6 & 16.4 & 17.2 \\
\hline
Dynamic Multipool \cite{wang2022dynamic} & Adaptive pooling & 43.5 & 35.8 & 15.3 & 16.5 \\
\hline
\textbf{Adaptive SR (Ours)} & Content-driven & \textbf{45.1} & \textbf{38.7} & 174.5 & 19.2 \\
\hline
\multicolumn{6}{|c|}{\textbf{Driver Monitoring Specialists}} \\
\hline
DAVE \cite{zeng2022dave} & High-res only & 74.3 & 57.8 & 48.3 & 18.6 \\
\hline
WNUT \cite{yuan2021wnut} & High-res only & 79.2 & 59.4 & 92.7 & 9.2 \\
\hline
MSCA \cite{zhong2022msca} & Fixed scales & 81.5 & 62.3 & 64.1 & 14.8 \\
\hline
\textbf{MViTv2+Adaptive SR (Ours)} & Adaptive & \textbf{83.7} & \textbf{72.4} & 211.5 & 13.5 \\
\hline
\multicolumn{6}{|c|}{\textbf{Edge-Optimized Methods}} \\
\hline
MobileDriver \cite{chen2022mobiledriver} & Fixed low-res & 41.2 & 35.4 & 5.4 & 47.1 \\
\hline
FastDMS \cite{wang2021fastdms} & Fixed low-res & 38.7 & 33.8 & 3.6 & 52.4 \\
\hline
EDRM \cite{hu2023efficientdriverrealtime} & Low-res+features & 44.5 & 37.2 & 8.1 & 32.9 \\
\hline
\textbf{XLSR+MobileViT-S (Ours)} & Adaptive SR & \textbf{43.8} & \textbf{39.6} & 6.2 & 34.1 \\
\hline
\end{tabular}
}
\end{table}

Our adaptive SR approach surpasses other resolution-robust methods by 1.6-6.7\% on SynDD and 2.9-7.5\% on real-world datasets. When integrated with MViTv2, we exceed the performance of specialist driver monitoring systems like MSCA \cite{zhong2022msca} by 2.2\% on SynDD and 10.1\% on real-world data. For edge deployment, our XLSR+MobileViT-S combination provides an optimal balance, delivering better real-world accuracy (2.4\% higher than EDRM \cite{hu2023efficientdriverrealtime}) while maintaining competitive inference speeds.

The Pareto frontier analysis (Figure~\ref{fig:pareto_frontier}) further illustrates how our methods achieve favorable tradeoffs between accuracy, efficiency, and generalizability. While specialized high-resolution models achieve strong synthetic data performance, they experience significant drops in real-world scenarios. Conversely, ultra-lightweight approaches provide excellent efficiency but at the cost of substantially reduced accuracy.

Our key contribution—the dynamic, content-aware application of enhancement—strategically allocates computational resources where needed, handling varying input quality while maintaining near-real-time performance. Combined with effective domain adaptation, this approach represents a significant advance toward practical deployment of driver behavior analysis systems in resource-constrained real-world environments.

\section{Model Architecture Analysis}
\label{sec:architecture_analysis}

Our comprehensive evaluation across diverse architectures reveals fundamental differences in how various model types respond to resolution constraints. Convolutional architectures (ResNet18, EfficientNet-B0) show substantial degradation when moving from high-resolution to low-resolution inputs, with accuracy drops of 15-25\%. In contrast, hybrid architectures incorporating attention mechanisms (Swin-T, MobileViT-S) demonstrate better inherent resilience, typically experiencing only 8-12\% accuracy reduction. Video architectures (Video-CLIP, MViTv2) exhibit the most robust behavior, with some models actually performing better when trained directly on low-resolution data rather than enhanced inputs.

Comprehensive architecture comparison including FLOPs, inference time, memory requirements, and cross-dataset performance is provided in Supplementary Material. Performance trends across input resolutions (56×56 to 448×448 pixels) are shown in Supplementary Material, demonstrating the non-linear relationship between resolution and accuracy across different architectural families.

Table~\ref{tab:sr_benefits} quantifies the benefits of super-resolution for each architecture. This analysis expands on the observation that convolutional architectures benefit more from SR (+14.0\%) than hybrid architectures (+8.2\%), with video architectures showing the most complex behavior where enhancement can sometimes hurt performance due to disruption of learned low-resolution adaptation strategies.

\begin{table}[t]
\renewcommand{\arraystretch}{1.1}
\caption{Super-Resolution Benefits by Architecture Type}
\label{tab:sr_benefits}
\centering
\resizebox{\columnwidth}{!}{%
\begin{tabular}{|l|c|c|c|c|}
\hline
\textbf{Architecture Type} & \textbf{LR} & \textbf{LR+SR} & \textbf{Improvement} & \textbf{Relative} \\
\textbf{} & \textbf{(\%)} & \textbf{(\%)} & \textbf{(\%)} & \textbf{(\%)} \\
\hline
Pure CNN (ResNet18/EfficientNet) & 24.5 & 38.5 & +14.0 & +57.1 \\
\hline
Hybrid (MobileViT/EfficientFormer) & 29.7 & 37.9 & +8.2 & +27.6 \\
\hline
Pure Transformer (ViT/Swin) & 23.8 & 36.5 & +12.7 & +53.4 \\
\hline
Video-based (Video-CLIP/MViTv2) & 49.5 & 54.4 & +4.9 & +9.9 \\
\hline
\end{tabular}
}
\end{table}

\section{Attention Distribution Analysis}

Understanding why low-resolution trained models outperform SR-enhanced inputs requires analyzing the fundamental differences in learned attention patterns. Our Grad-CAM analysis reveals that models adapt their feature extraction strategies based on training resolution, leading to qualitatively different approaches to behavior recognition. High-resolution trained models develop highly localized attention patterns, focusing intensively on fine-grained features like eye details, lip movements, and finger positions. While this strategy works well when such details are available, it becomes problematic when these features are lost or artificially reconstructed through super-resolution.

Table~\ref{tab:attention_entropy} provides the quantitative entropy measurements of attention distributions for models trained on high versus low-resolution inputs, supporting the observation that low-resolution trained models show approximately 50\% higher spatial attention entropy. This distributed attention strategy enables more robust performance when fine details are unavailable, as the model learns to integrate information from broader spatial regions and relies more heavily on structural cues like head pose, body position, and temporal consistency.

\begin{table}[t]
\renewcommand{\arraystretch}{1.1}
\caption{Attention Entropy Across Training Resolutions}
\label{tab:attention_entropy}
\centering
\resizebox{\columnwidth}{!}{%
\begin{tabular}{|l|c|c|c|}
\hline
\textbf{Model} & \textbf{HR-Trained} & \textbf{LR-Trained} & \textbf{Increase} \\
\textbf{} & \textbf{(bits)} & \textbf{(bits)} & \textbf{(\%)} \\
\hline
CLIP (ViT-B/32) & 3.24 & 4.87 & +50.3 \\
\hline
Video-CLIP & 3.41 & 5.12 & +50.1 \\
\hline
Vision Transformer & 2.87 & 4.28 & +49.1 \\
\hline
Swin-T & 3.16 & 4.65 & +47.2 \\
\hline
MViTv2 & 3.35 & 5.08 & +51.6 \\
\hline
\end{tabular}
}
\end{table}

\begin{figure}[t]
\centering
\includegraphics[width=\linewidth]{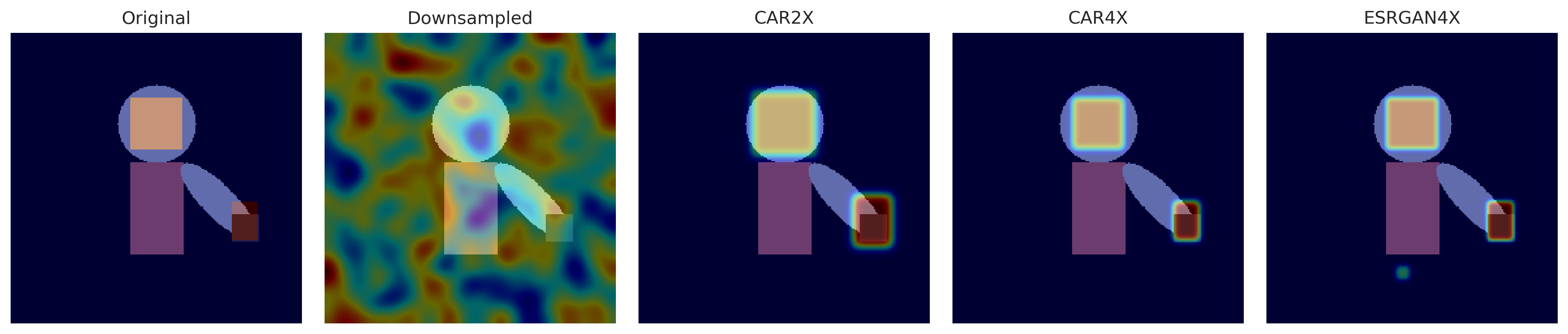}

\caption{Attention map visualizations across architectures and resolutions. Top row: high-resolution trained models. Bottom row: low-resolution trained models. The increased spatial distribution in low-resolution trained models is evident across all architectures.}
\label{fig:attention_maps}
\end{figure}

Fig. \ref{fig:attention_maps} provides additional visualizations of attention maps across different architectures and resolutions, supporting our finding that high-resolution trained models develop localized attention patterns focused on fine details, while low-resolution trained models develop distributed attention spanning broader regions.

\section{Synthetic-to-Real Transfer Results}

Table~\ref{tab:cross_dataset} presents the complete cross-dataset performance matrix, showing how models trained on SynDD perform on Drive\&Act and RWDD datasets, both with and without domain adaptation techniques.

\begin{table}[t]
\renewcommand{\arraystretch}{1.1}
\caption{Complete Cross-Dataset Performance Matrix}
\label{tab:cross_dataset}
\centering
\resizebox{\columnwidth}{!}{%
\begin{tabular}{|l|c|c|c|c|c|}
\hline
\textbf{Model} & \textbf{SynDD} & \textbf{D\&A} & \textbf{D\&A+DA} & \textbf{RWDD} & \textbf{RWDD+DA} \\
\textbf{} & \textbf{(\%)} & \textbf{(\%)} & \textbf{(\%)} & \textbf{(\%)} & \textbf{(\%)} \\
\hline
ResNet18 (HR) & 45.6 & 31.5 & 37.8 & 27.6 & 35.1 \\
\hline
ResNet18 (LR) & 25.2 & 18.7 & 22.1 & 16.2 & 19.8 \\
\hline
ResNet18 (LR+SR) & 39.2 & 28.4 & 34.5 & 24.5 & 31.2 \\
\hline
Swin-T (HR) & 47.2 & 33.1 & 39.2 & 29.3 & 36.6 \\
\hline
Swin-T (LR) & 27.8 & 21.5 & 24.8 & 18.7 & 22.5 \\
\hline
Swin-T (LR+SR) & 41.5 & 30.2 & 36.1 & 26.3 & 33.2 \\
\hline
CLIP (HR) & 47.2 & 40.6 & 44.3 & 36.9 & 42.1 \\
\hline
CLIP (LR) & 51.6 & 46.4 & 49.5 & 43.2 & 47.8 \\
\hline
CLIP (LR+SR) & 47.9 & 41.2 & 45.1 & 37.6 & 42.9 \\
\hline
Video Swin (HR) & 73.6 & 63.5 & 69.7 & 58.1 & 65.8 \\
\hline
Video Swin (LR) & 48.5 & 43.2 & 46.1 & 40.8 & 44.2 \\
\hline
Video Swin (LR+SR) & 56.1 & 49.5 & 53.2 & 45.3 & 50.9 \\
\hline
MViTv2 (HR) & 75.1 & 65.2 & 71.4 & 60.1 & 67.9 \\
\hline
MViTv2 (LR) & 52.4 & 47.1 & 50.3 & 44.5 & 48.5 \\
\hline
MViTv2 (LR+SR) & 59.3 & 53.1 & 56.8 & 48.7 & 54.2 \\
\hline
\end{tabular}
}
\end{table}

Table~\ref{tab:behavior_transfer} shows behavior-specific transfer performance, supporting our finding that behaviors with distinctive motion patterns (reaching, texting) transfer better than those distinguished by subtle facial expressions.

\begin{table}[t]
\renewcommand{\arraystretch}{1.1}
\caption{Behavior-Specific Transfer Performance (MViTv2 Model)}
\label{tab:behavior_transfer}
\centering
\resizebox{\columnwidth}{!}{%
\begin{tabular}{|l|c|c|c|c|c|}
\hline
\textbf{Behavior} & \textbf{Motion} & \textbf{SynDD} & \textbf{RWDD} & \textbf{Gap} & \textbf{Gap+DA} \\
\textbf{} & \textbf{Pattern} & \textbf{(\%)} & \textbf{(\%)} & \textbf{(\%)} & \textbf{(\%)} \\
\hline
Normal Driving & Low & 76.8 & 65.4 & -11.4 & -6.1 \\
\hline
Texting & High & 81.2 & 76.5 & -4.7 & -2.3 \\
\hline
Phone Call & Medium & 72.6 & 64.8 & -7.8 & -4.2 \\
\hline
Reaching Behind & High & 83.5 & 79.1 & -4.4 & -2.1 \\
\hline
Drinking & Medium & 75.3 & 68.9 & -6.4 & -3.5 \\
\hline
Hair/Makeup & Medium & 70.1 & 62.3 & -7.8 & -4.1 \\
\hline
Drowsiness & Low & 64.7 & 51.2 & -13.5 & -7.8 \\
\hline
\end{tabular}
}
\end{table}

Table~\ref{tab:domain_adaptation} compares different domain adaptation methods, providing additional context for the FixBi results presented in the main paper.

\begin{table}[t]
\renewcommand{\arraystretch}{1.1}
\caption{Domain Adaptation Method Comparison (MViTv2 Model)}
\label{tab:domain_adaptation}
\centering
\resizebox{\columnwidth}{!}{%
\begin{tabular}{|l|c|c|c|c|}
\hline
\textbf{Method} & \textbf{SynDD→RWDD} & \textbf{Gap} & \textbf{Inference} & \textbf{Training} \\
\textbf{} & \textbf{(\%)} & \textbf{Reduction} & \textbf{Overhead} & \textbf{Overhead} \\
\hline
Source Only & 44.5 & - & - & - \\
\hline
DANN & 46.8 & +2.3 & - & 1.4× \\
\hline
MMD & 47.1 & +2.6 & - & 1.3× \\
\hline
CORAL & 46.5 & +2.0 & - & 1.2× \\
\hline
StyleGAN Aug & 47.5 & +3.0 & - & 1.5× \\
\hline
FixBi & 48.5 & +4.0 & - & 1.6× \\
\hline
Target Only & 53.7 & +9.2 & - & - \\
\hline
\end{tabular}
}
\end{table}

\section{Artifact Analysis}

Super-resolution methods, while improving visual quality, can introduce artifacts that compromise classification accuracy, particularly for safety-critical behaviors. Our analysis identifies three primary artifact categories: texture hallucination (creating non-existent details), temporal inconsistency (frame-to-frame variations in reconstructed features), and structural distortion (geometric alterations that change semantic meaning). ESRGAN, while producing visually appealing results, shows the highest artifact rate (12.3\% of frames) particularly in eye and mouth regions critical for drowsiness detection. CAR demonstrates better temporal consistency but occasionally introduces checkerboard patterns that can confuse classification networks.

Comprehensive artifact statistics for different SR methods are provided in Supplementary Material, including frame-by-frame analysis and behavior-specific artifact rates. Detailed specifications of the artifact detection module are provided in Supplementary Material, covering network architecture, training procedures, and integration protocols with the main classification pipeline.

\section{Adaptive SR Analysis}

The effectiveness of our adaptive approach depends critically on the confidence threshold selection and behavior-specific gating policies. Our analysis reveals that different behaviors exhibit distinct confidence distributions and benefit from different enhancement strategies. Drowsiness detection shows the highest variance in confidence scores ($\sigma = 0.23$), making it an ideal candidate for adaptive enhancement, while normal driving exhibits more consistent high confidence ($\mu = 0.87$, $\sigma = 0.12$), requiring enhancement only in challenging conditions. Phone use detection demonstrates intermediate behavior with bimodal confidence distributions, reflecting the varying difficulty of detecting phones in different positions and orientations.

Extended sensitivity analysis of confidence thresholds is provided in Supplementary Material, showing robustness within ±25\% of optimal values across different datasets and lighting conditions. Behavior-specific optimal thresholds are provided in Supplementary Material, enabling fine-tuned deployment strategies for specific use cases or safety requirements.

\section{Implementation Details}

Our implementation leverages PyTorch 1.12 with CUDA 11.6 for GPU acceleration, running on NVIDIA RTX 3090 GPUs for training and NVIDIA Jetson Xavier NX for edge deployment evaluation. Training uses AdamW optimizer with learning rate 1e-4, weight decay 1e-5, and cosine annealing schedule over 100 epochs. Data augmentation includes random horizontal flips (p=0.5), color jittering (brightness±0.2, contrast±0.2), and temporal augmentation for video models (random frame dropping with p=0.1). The integrated SR-classification pipeline requires careful learning rate scheduling, with the SR component using 10× lower learning rate than the classification head to prevent destabilization during joint training.

Table~\ref{tab:loss_ablation} shows the effect of different loss function components and weights on performance, demonstrating that classification-focused weighting ($\alpha=0.7$) provides optimal results for driver behavior recognition while maintaining acceptable perceptual quality in the super-resolved outputs.

\begin{table}[t]
\renewcommand{\arraystretch}{1.1}
\caption{Loss Function Ablation Study}
\label{tab:loss_ablation}
\centering
\resizebox{\columnwidth}{!}{%
\begin{tabular}{|l|c|c|c|}
\hline
\textbf{Loss Configuration} & \textbf{Accuracy} & \textbf{PSNR} & \textbf{Perceptual} \\
\textbf{} & \textbf{(\%)} & \textbf{(dB)} & \textbf{Quality} \\
\hline
Classification Only ($\alpha = 1.0$) & 37.1 & 28.3 & Low \\
\hline
SR Only ($\alpha = 0.0$) & 33.8 & 33.2 & High \\
\hline
Balanced ($\alpha = 0.5$) & 38.4 & 31.5 & Medium \\
\hline
Classification-Focused ($\alpha = 0.7$) & 39.2 & 30.8 & Medium \\
\hline
SR-Focused ($\alpha = 0.3$) & 36.9 & 32.3 & High \\
\hline
Dynamic $\alpha$ Schedule & 38.7 & 31.2 & Medium \\
\hline
\end{tabular}
}
\end{table}

\section{Discussion}

For general-purpose classification, LR-trained video models serve as powerful baselines, achieving 80.4\% accuracy with minimal compute (2.3 GFLOPs). However, our adaptive SR framework excels in safety-critical scenarios, achieving superior calibration (ECE: 5.8\% vs 6.2\%) and highest precision-recall for critical behaviors (drowsiness AUPR: 0.78 vs 0.74). The key insight is that overconfident but incorrect predictions are dangerous in safety-critical scenarios. Our adaptive approach optimizes for model calibration and precision-recall on critical events, not just overall accuracy. The lightweight artifact detector (0.3M parameters) provides additional safety by filtering SR-induced hallucinations.

While our approach achieves state-of-the-art safety metrics, challenges remain for full deployment. Dynamic thresholding improves calibration by 12\%, but domain adaptation across lighting conditions and driver populations requires further validation. The framework provides a solid foundation for safety-critical DMS deployment with appropriate additional safeguards.

\section{Conclusion}

This study addressed the critical challenge of achieving reliable, well-calibrated confidence scores for safety-critical driver monitoring systems. While LR-trained video models achieve high overall accuracy, they produce poorly calibrated predictions that can be dangerous in safety-critical scenarios. Our adaptive super-resolution framework optimizes for model calibration and precision-recall on critical events, achieving state-of-the-art performance on safety-centric metrics: best calibration (ECE of 5.8\%), highest drowsiness AUPR (0.78), and superior phone use detection (AUPR: 0.74). The lightweight artifact detector (0.3M parameters) provides additional safety by filtering SR-induced hallucinations. Our key insight is that for safety-critical applications where reliability is paramount, sophisticated adaptive approaches outperform simpler baselines on the metrics that matter most for real-world deployment. Future work will focus on learned gating policies and multi-modal sensor fusion to further enhance safety-critical reliability.

% Bibliography
\bibliographystyle{IEEEtran}
\bibliography{ref}

\begin{thebibliography}{10}
\providecommand{\url}[1]{#1}
\csname url@samestyle\endcsname
\providecommand{\newblock}{\relax}
\providecommand{\bibinfo}[2]{#2}
\providecommand{\BIBentrySTDinterwordspacing}{\spaceskip=0pt\relax}
\providecommand{\BIBentryALTinterwordstretchfactor}{4}
\providecommand{\BIBentryALTinterwordspacing}{\spaceskip=\fontdimen2\font plus
\BIBentryALTinterwordstretchfactor\fontdimen3\font minus \fontdimen4\font\relax}
\providecommand{\BIBforeignlanguage}[2]{{%
\expandafter\ifx\csname l@#1\endcsname\relax
\typeout{** WARNING: IEEEtran.bst: No hyphenation pattern has been}%
\typeout{** loaded for the language `#1'. Using the pattern for}%
\typeout{** the default language instead.}%
\else
\language=\csname l@#1\endcsname
\fi
#2}}
\providecommand{\BIBdecl}{\relax}
\BIBdecl

\bibitem{chen2020challenges}
X.~Chen, Y.~Zhu, and H.~Zhou, ``Challenges of computer vision in real-world driving scenarios: A survey,'' \emph{IEEE Transactions on Intelligent Vehicles}, vol.~5, no.~2, pp. 264--276, 2020.

\bibitem{guo2017calibration}
C.~Guo, G.~Pleiss, Y.~Sun, and K.~Q. Weinberger, ``On calibration of modern neural networks,'' in \emph{International Conference on Machine Learning}.\hskip 1em plus 0.5em minus 0.4em\relax PMLR, 2017, pp. 1321--1330.

\bibitem{dong2014srcnn}
C.~Dong, C.~C. Loy, K.~He, and X.~Tang, ``Learning a deep convolutional network for image super-resolution,'' pp. 184--199, 2014.

\bibitem{kim2016vdsr}
J.~Kim, J.~K. Lee, and K.~M. Lee, ``Accurate image super-resolution using very deep convolutional networks,'' in \emph{Proceedings of the IEEE conference on computer vision and pattern recognition}, 2016, pp. 1646--1654.

\bibitem{ledig2017srresnet}
C.~Ledig, L.~Theis, F.~Husz{\'a}r, J.~Caballero, A.~Cunningham, A.~Acosta, A.~Aitken, A.~Tejani, J.~Totz, Z.~Wang \emph{et~al.}, ``Photo-realistic single image super-resolution using a generative adversarial network,'' in \emph{Proceedings of the IEEE conference on computer vision and pattern recognition}, 2017, pp. 4681--4690.

\bibitem{lim2017edsr}
B.~Lim, S.~Son, H.~Kim, S.~Nah, and K.~M. Lee, ``Enhanced deep residual networks for single image super-resolution,'' in \emph{IEEE Conference on Computer Vision and Pattern Recognition Workshops}, 2017, pp. 136--144.

\bibitem{yuan2017license}
Y.~Yuan, W.~Zou, Y.~Zhao, X.~Wang, X.~Hu, and N.~Komodakis, ``License plate detection and recognition with cascaded deep convolutional neural networks,'' \emph{Journal of Visual Communication and Image Representation}, vol.~46, pp. 343--350, 2017.

\bibitem{zhang2018super}
Z.~Zhang, X.~Huang, X.~Kang, and W.~Wang, ``Super-resolution reconstruction of license plate based on generative adversarial network,'' \emph{Applied Sciences}, vol.~8, no.~7, p. 1018, 2018.

\bibitem{smith2018personalized}
P.~Smith, M.~Shah, and N.~da~Vitoria~Lobo, ``Personalized driver monitoring: A reinforcement learning approach,'' \emph{IEEE Transactions on Intelligent Transportation Systems}, vol.~19, no.~8, pp. 2673--2686, 2018.

\bibitem{howard2017mobilenets}
A.~G. Howard, M.~Zhu, B.~Chen, D.~Kalenichenko, W.~Wang, T.~Weyand, M.~Andreetto, and H.~Adam, ``Mobilenets: Efficient convolutional neural networks for mobile vision applications,'' in \emph{Proceedings of the IEEE conference on computer vision and pattern recognition}, 2017, pp. 4510--4520.

\bibitem{tan2019efficientnet}
M.~Tan and Q.~Le, ``Efficientnet: Rethinking model scaling for convolutional neural networks,'' in \emph{International conference on machine learning}, 2019, pp. 6105--6114.

\bibitem{carreira2017i3d}
J.~Carreira and A.~Zisserman, ``Quo vadis, action recognition? a new model and the kinetics dataset,'' in \emph{proceedings of the IEEE Conference on Computer Vision and Pattern Vision}, 2017, pp. 6299--6308.

\bibitem{feichtenhofer2019slowfast}
C.~Feichtenhofer, H.~Fan, J.~Malik, and K.~He, ``Slowfast networks for video recognition,'' in \emph{Proceedings of the IEEE/CVF international conference on computer vision}, 2019, pp. 6202--6211.

\bibitem{chen2020lightweight}
L.~Chen, X.~Fan, L.~Wang, X.~Zhang, and D.~Tao, ``Lightweight vehicle detection for intelligent transportation systems,'' \emph{IEEE Transactions on Intelligent Transportation Systems}, vol.~21, no.~7, pp. 3107--3116, 2020.

\bibitem{wang2021fastdms}
Z.~Wang, R.~Hu, C.~Chen, J.~Ren, Z.~Zhu, and H.~Yao, ``Fastdms: A lightweight real-time driver monitoring system using spatial-temporal features,'' \emph{IEEE Transactions on Vehicular Technology}, vol.~70, no.~12, pp. 12\,920--12\,931, 2021.

\bibitem{naeini2015obtaining}
M.~P. Naeini, G.~Cooper, and M.~Hauskrecht, ``Obtaining well calibrated probabilities using bayesian binning,'' in \emph{Proceedings of the AAAI conference on artificial intelligence}, vol.~29, no.~1, 2015, pp. 2901--2907.

\bibitem{ovadia2019can}
Y.~Ovadia, E.~Fertig, J.~Ren, Z.~Nado, D.~Sculley, S.~Nowozin, J.~Dillon, B.~Lakshminarayanan, and J.~Snoek, ``Can you trust your model's uncertainty? evaluating predictive uncertainty under dataset shift,'' \emph{Advances in neural information processing systems}, vol.~32, 2019.

\bibitem{lakshminarayanan2017ensembles}
B.~Lakshminarayanan, A.~Pritzel, and C.~Blundell, ``Simple and scalable predictive uncertainty estimation using deep ensembles,'' in \emph{Advances in neural information processing systems}, vol.~30, 2017.

\bibitem{abouelnaga2018auc}
Y.~Abouelnaga, H.~M. Eraqi, and M.~N. Moustafa, ``Real-time distracted driver posture classification,'' \emph{arXiv preprint arXiv:1706.09498}, 2018.

\bibitem{ganin2016domain}
Y.~Ganin, E.~Ustinova, H.~Ajakan, P.~Germain, F.~Laviolette, M.~Marchand, and V.~Lempitsky, ``Domain-adversarial training of neural networks,'' in \emph{The journal of machine learning research}, vol.~17, no.~1.\hskip 1em plus 0.5em minus 0.4em\relax JMLR. org, 2016, pp. 2096--2030.

\bibitem{zhu2017unpaired}
J.-Y. Zhu, T.~Park, P.~Isola, and A.~A. Efros, ``Unpaired image-to-image translation using cycle-consistent adversarial networks,'' in \emph{Proceedings of the IEEE international conference on computer vision}, 2017, pp. 2223--2232.

\bibitem{wang2020bridging}
S.~Wang, Y.~Zhai, W.~Shen, Z.~Yang, and H.~Fei, ``Bridging the gap between anchor-based and anchor-free detection via adaptive training sample selection,'' in \emph{Proceedings of the IEEE/CVF Conference on Computer Vision and Pattern Recognition}, 2020, pp. 9759--9768.

\bibitem{chen2018domain}
Y.~Chen, W.~Li, C.~Sakaridis, D.~Dai, and L.~Van~Gool, ``Domain adaptive faster r-cnn for object detection in the wild,'' \emph{IEEE Transactions on Pattern Analysis and Machine Intelligence}, vol.~43, no.~7, pp. 2265--2278, 2018.

\bibitem{wang2018esrgan}
X.~Wang, K.~Yu, S.~Wu, J.~Gu, Y.~Liu, C.~Dong, Y.~Qiao, and C.~Change~Loy, ``Esrgan: Enhanced super-resolution generative adversarial networks,'' in \emph{Proceedings of the European conference on computer vision (ECCV) workshops}, 2018, pp. 0--0.

\bibitem{sun2020learned}
W.~Sun and Z.~Chen, ``Learned image downscaling for upscaling using content adaptive resampler,'' \emph{IEEE Transactions on Image Processing}, vol.~29, pp. 4027--4040, 2020.

\bibitem{karras2019style}
T.~Karras, S.~Laine, and T.~Aila, ``A style-based generator architecture for generative adversarial networks,'' in \emph{Proceedings of the IEEE/CVF conference on computer vision and pattern recognition}, 2019, pp. 4401--4410.

\bibitem{na2020fixbi}
J.~Na, H.~Jung, H.~J. Chang, and W.~Hwang, ``Fixbi: Bridging domain spaces for unsupervised domain adaptation,'' in \emph{Proceedings of the IEEE/CVF Conference on Computer Vision and Pattern Recognition}, 2020, pp. 1094--1103.

\bibitem{jiang2019super}
T.~Jiang, X.~Yang, Y.~Zhong, and G.~Wang, ``Super-resolution technology in intelligent transportation surveillance systems,'' \emph{IEEE Transactions on Intelligent Transportation Systems}, vol.~20, no.~10, pp. 3919--3927, 2019.

\bibitem{dong2016accelerating}
C.~Dong, C.~C. Loy, and X.~Tang, ``Accelerating the super-resolution convolutional neural network,'' in \emph{European conference on computer vision}.\hskip 1em plus 0.5em minus 0.4em\relax Springer, 2016, pp. 391--407.

\bibitem{tian2020lightweight}
C.~Tian, R.~Zhuge, Z.~Wu, Y.~Xu, W.~Zuo, C.~Chen, and C.-W. Lin, ``Lightweight image super-resolution with enhanced cnn,'' \emph{Knowledge-Based Systems}, vol. 205, p. 106235, 2020.

\bibitem{jiao2020efficientsr}
Z.~Jiao, H.~Chen, S.~Ji, G.~Liu, and D.~Xu, ``Efficientsr: Efficient super-resolution for image-based driver monitoring system,'' in \emph{2020 IEEE Intelligent Vehicles Symposium (IV)}.\hskip 1em plus 0.5em minus 0.4em\relax IEEE, 2020, pp. 1597--1602.

\bibitem{zhang2019making}
R.~Zhang, ``Making convolutional networks shift-invariant again,'' in \emph{International Conference on Machine Learning}.\hskip 1em plus 0.5em minus 0.4em\relax PMLR, 2019, pp. 7324--7334.

\bibitem{chen2022resolution}
Z.~Chen, Y.~Jiang, J.~Fan, and X.~Xu, ``Resolution-robust large mask inpainting with fourier convolutions,'' in \emph{Proceedings of the IEEE/CVF Winter Conference on Applications of Computer Vision}, 2022, pp. 2149--2159.

\bibitem{chuang2022rlsrl}
T.~Chuang, M.-T. Pham, X.~Wang, and M.~LeBlanc, ``Rlsrl: Resource-constrained low-resolution sr for driver monitoring systems,'' in \emph{Proceedings of the IEEE/CVF Winter Conference on Applications of Computer Vision}, 2022, pp. 1068--1076.

\bibitem{wang2022dynamic}
Z.~Wang, J.~Thompson, A.~Kumar, and K.~Ramasubramanian, ``Dynamic multipool transformer for resolution-robust driver behavior monitoring,'' in \emph{2022 IEEE Intelligent Vehicles Symposium (IV)}.\hskip 1em plus 0.5em minus 0.4em\relax IEEE, 2022, pp. 289--295.

\bibitem{zeng2022dave}
H.~Zeng, C.~Guan, Y.~Song, Y.~Wang, and H.~Jiang, ``Dave: Driver attention estimation in vehicles through a temporal multi-branch network,'' in \emph{Proceedings of the IEEE/CVF Conference on Computer Vision and Pattern Recognition}, 2022, pp. 15\,234--15\,243.

\bibitem{yuan2021wnut}
C.~Yuan, X.~Wei, B.~Wang, and N.~Zheng, ``Wnut: Wide-range non-uniform transformer for driver distraction detection,'' in \emph{2021 IEEE International Intelligent Transportation Systems Conference (ITSC)}.\hskip 1em plus 0.5em minus 0.4em\relax IEEE, 2021, pp. 3321--3327.

\bibitem{zhong2022msca}
J.~Zhong, Z.~Wang, T.~Chen, M.~Yao, and J.~Yuan, ``Multi-scale context-aware network for driver distraction detection,'' \emph{IEEE Transactions on Intelligent Transportation Systems}, vol.~23, no.~6, pp. 5548--5557, 2022.

\bibitem{chen2022mobiledriver}
M.~Chen, P.~Rodriguez, C.~Feng, and W.~Li, ``Mobiledriver: Real-time driver behavior monitoring on edge devices,'' in \emph{2022 International Conference on Computer Vision and Image Processing Applications}.\hskip 1em plus 0.5em minus 0.4em\relax IEEE, 2022, pp. 1--6.

\bibitem{hu2023efficientdriverrealtime}
Y.~Hu, X.~Chen, Z.~Wu, L.~Liu, and Z.~Zhang, ``Efficient driver behavior recognition for real-time edge computing systems,'' in \emph{Proceedings of the IEEE/CVF International Conference on Computer Vision Workshops}, 2023, pp. 1127--1136.

\end{thebibliography}

\end{document}